\pdfoutput=1
\documentclass[11pt]{article}
\usepackage[final]{acl}

\usepackage{times}
\usepackage{latexsym}
\usepackage[T1]{fontenc}
\usepackage[utf8]{inputenc}
\usepackage{microtype}
\usepackage{inconsolata}
\usepackage{comment}
\usepackage{booktabs}  
\usepackage{colortbl} 
\usepackage{multirow} 
\usepackage{amssymb}
\usepackage{amsmath}
\usepackage{pdfpages}  
\usepackage{subcaption} 
\usepackage{graphicx}  
\usepackage{float}
\usepackage{hyperref}
\usepackage[whole]{bxcjkjatype}
\usepackage{amsmath}
\usepackage{circledsteps}
\definecolor{LightGray}{gray}{0.9}
\usepackage{makecell}
\usepackage{tabularx}
\usepackage[utf8]{inputenc}
\usepackage[most]{tcolorbox}
\definecolor{LightGray}{rgb}{0.97,0.97,0.97}
\usepackage[verb]{bxghost}

\usepackage{subcaption} 
\usepackage{listings}

\title{Investigating Learner-Aware Design \\ of LLM-Generated Educational Feedback}

\author{
Momoka Furuhashi${}^{1,2}$
Kouta Nakayama${}^{2}$ 
Noboru Kawai${}^{3}$ \\
\bfseries
Takashi Kodama${}^{2}$ 
Saku Sugawara${}^{4,5}$ 
Kyosuke Takami${}^{3}$ \\
${}^{1}$Tohoku University \hspace{1em}
${}^{2}$Research and Development Center for Large Language Models,\\
National Institute of Informatics \hspace{1em}
${}^{3}$Osaka Kyoiku University\\
${}^{4}$National Institute of Informatics \hspace{1em}
${}^{5}$University of Tokyo\\
\texttt{furuhashi.momoka.p4@dc.tohoku.ac.jp} \hspace{1em}
\texttt{\{nakayama,tkodama,saku\}@nii.ac.jp}\\
\texttt{t-kawain@e.osakamanabi.jp} \hspace{1em}
\texttt{takami-k75@cc.osaka-kyoiku.ac.jp}
}

\begin{document}
\maketitle
\begin{abstract}
Although large language models (LLMs) show promise for generating educational feedback, it remains unclear how feedback should be designed (e.g., tone and coverage) to support answer revision and learner evaluations across learner profiles.
We define six feedback designs for multiple-choice biology questions, including a baseline design and five variants with additional feedback elements, and conduct an empirical study with 321 high school students.
We evaluate feedback using immediate revision performance and six subjective evaluation criteria, and analyze differences in subjective evaluations across learner profiles based on personality traits.
Our results show that presenting task-relevant information clearly is associated with better immediate revision performance and is favorably evaluated across learner profiles, while we observe descriptive differences in evaluation patterns, particularly for informational novelty and affective framing.
These findings support further investigation of personalized LLM feedback design.
\end{abstract}

\section{Introduction}
\label{sec:introduction}
\begin{figure*}[t]
\includegraphics[width=\textwidth]{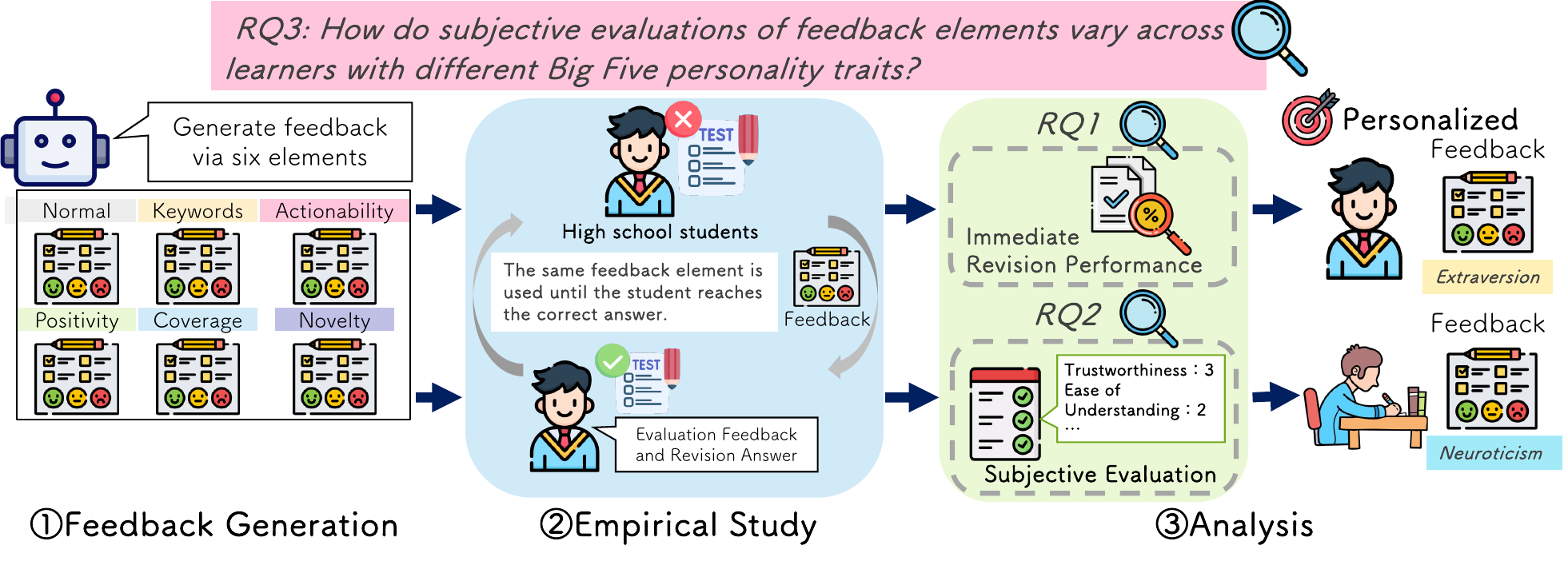}
\caption{Overview of this study.
First, we define six feedback elements related to feedback content and generate feedback using GPT-5.
Next, based on data from 321 participants, we analyze immediate revision performance and subjective evaluations across six criteria.
Finally, we examine which feedback elements are more favorably received across learners with different Big Five personality traits.}
\label{fig:figure1}
\end{figure*}

Large language models (LLMs) have enabled the large-scale generation of diverse forms of educational feedback and have accelerated the development of automated learning support systems~\cite{kasneci2023chatgpt,stamper2024enhancing,chu-etal-2025-llm}.
Unlike conventional template-based approaches~\cite{lai-chang-2019-tellmewhy}, LLMs can flexibly generate feedback with different educational designs, such as explanation level and tone.
Feedback helps learners bridge the gap between their understanding and correct answers~\cite{Hattie2007,wisniewski2020power}, and must be understandable and trustworthy enough to guide learners toward productive answer revision.

Previous studies on human-designed feedback also have shown that feedback effectiveness differs across learner profiles such as prior knowledge and personality traits ~\cite{Aleven2006TowardMT,Bidjerano2007TheRB,poropat2009meta}.Therefore, personalized feedback generation requires consideration of learner profiles as well as feedback content.
To design effective LLM-generated feedback, we need to identify which feedback designs positively influence learner behavior and perception through empirical studies.
Although LLMs can generate adaptive feedback, the effectiveness of different designs for learners remains underexplored.

Prior work has compared human-written and LLM-generated feedback~\cite{Escalante2023AIgeneratedFO,steiss2024comparing,asadi2025impact} or evaluated generated feedback using LLM-based evaluation~\cite{qian2025deanllmtutorsexploring,chu2026feedevalpedagogicallyalignedevaluation}.
However, these studies do not systematically investigate how feedback designs influence learners, and it remains unclear whether LLM-based evaluations align with learner perceptions and behaviors. Although~\citet{borges-etal-2024-teach} identify educational elements that may influence the quality of LLM-generated feedback, such as coverage and tone, their effects on learner behavior and feedback perception have not been systematically investigated.

To address the limited consideration of learner profiles and the unclear effects of feedback elements in LLM-generated feedback, we investigate the following three research questions through the three-step workflow in Figure~\ref{fig:figure1}.
\begin{itemize}
\item \textit{RQ1: Which feedback elements most effectively improve revision performance?}
\item \textit{RQ2: Which feedback elements lead to more favorable subjective evaluations?}
\item \textit{RQ3: How do subjective evaluations of feedback elements vary across learners with different Big Five personality traits?
}
\end{itemize}

First, we define six feedback elements related to content, such as information coverage and tone, based on~\citet{borges-etal-2024-teach} and generate feedback for multiple-choice biology questions using gpt-5-2025-08-07~\cite{GPT-5}.
Next, through an empirical experiment with 321 first-year high school students, we evaluate feedback effects on immediate revision performance and six subjective evaluation criteria, such as trustworthiness and ease of understanding (\textit{RQ1, RQ2}).
Finally, we analyze subjective evaluations across learners with different Big Five personality traits (\textit{RQ3}).

We find that feedback presenting task-relevant points clearly is associated with better immediate revision.
Subjective evaluation results further show that learners tend to favor feedback that is practical, understandable, and easy to apply.
We also observe differences in subjective evaluation patterns across learner profiles.
Rather than indicating fixed preferences for particular feedback elements, these results suggest that learner characteristics may be associated with how feedback is perceived.
These findings are broadly consistent with prior work~\cite{komarraju2005relationship,Hattie2007,komarraju2009role,wisniewski2020power} and provide empirical support for further investigating learner-aware personalization of LLM-generated feedback.

The contributions of this study are as follows:
\begin{itemize}
    \item We define six controllable educational elements for LLM-generated feedback and empirically examine their effects on  revision performance and subjective evaluations through an experiment with 321 students.
    \item We provide empirical evidence that feedback evaluation patterns differ across learner profiles, motivating further investigation of learner-aware LLM feedback design.
\end{itemize}

\section{Related Work}
\label{sec:related_work}
\subsection{Feedback}
\label{ssec:feedback}
Feedback helps learners bridge the gap between their current state and learning goals, and can take various forms, ranging from simple correct/ incorrect indicators (e.g., checkmarks) to elaborated explanations that provide written guidance~\cite{Hattie2007,wisniewski2020power}.
These studies conceptualize feedback as a multi-dimensional construct, where explanations differ in whether they directly provide correct answers or encourage learners to derive them, as well as in content-related elements (e.g., level of elaboration and explanatory detail) and delivery-related criteria (e.g., timing and modality).
In this study, we focus on content-related elements that determine the granularity and informational richness of feedback.
We adopt an indirect feedback approach that encourages learners to derive answers.

\subsection{Feedback Generation}
\label{ssec:feedback_generation}
Prior to the emergence of LLMs, intelligent tutoring systems (ITSs) extensively investigated feedback generation using rule-based and model-based approaches, enabling adaptive and personalized instructional support based on learners' responses and problem-solving states~\cite{VanLehn2011TheRE,Kochmar2020AutomatedPF,villegas2025adaptive}.
Building on this foundation, recent studies have begun integrating LLM-generated feedback into classroom settings and examining its effects on learning outcomes and motivation~\cite{10260740,kinder2025effects}.
However, most evaluations of LLM-generated feedback rely either on coarse-grained comparisons with human-written feedback~\cite{Escalante2023AIgeneratedFO,steiss2024comparing} or automatic evaluation~\cite{qian2025deanllmtutorsexploring,chu2026feedevalpedagogicallyalignedevaluation}, neither of which captures how specific educational designs influence learners' perception and use of feedback.
Although \citet{borges-etal-2024-teach} identify several such elements, such as coverage and tone, their effects on learner behavior and feedback perception remain empirically underexplored.
Moreover, while prior educational studies indicate that feedback effectiveness varies across learner profiles~\cite{Aleven2006TowardMT,Bidjerano2007TheRB,poropat2009meta}, such individual differences have received limited attention in LLM-generated feedback research.
We investigate how different feedback elements influence learners' immediate revision performance and subjective evaluations, and how these effects vary across learners with different Big Five personality traits.

\subsection{Big Five}
Research on individual differences has long examined variability in learners' behavior and cognition, and the Big Five personality trait model (Openness, Conscientiousness, Extraversion, Agreeableness, and Neuroticism) is widely adopted as a framework for describing such differences~\cite{mccrae1992introduction}.
In educational research, these traits are used to characterize learners and have been linked to motivation, academic achievement, and preferences for instructional support and feedback~\cite{Aleven2006TowardMT,Bidjerano2007TheRB,poropat2009meta,takami2023personality}.
In Natural Language Processing, there is a growing interest in incorporating the Big Five framework into LLMs to represent user personas and control generated outputs. 
Recent studies have incorporated the Big Five framework into LLMs to represent user personas and control generated outputs~\cite{jiang-etal-2024-personallm,liu-etal-2024-personality}.
These studies simulate diverse learner profiles and examine personality-conditioned generation.
However, how learner personality shapes the perception and acceptance of generated feedback remains underexplored~\cite{Sharma2025}. Therefore, we analyze the relationship between learners' Big Five personality traits and their subjective evaluations of LLM-generated feedback.

\section{Feedback}
\label{sec:method}

\subsection{Feedback Elements}
We define six feedback elements: a manually designed \textit{Normal} condition and five educational elements based on~\citet{borges-etal-2024-teach}.

Although their research identifies ten elements for educational feedback design, we exclude \textit{Mode}, which represents multimodal feedback since our study focuses on text-based feedback.
Using the remaining nine elements and their sub-elements, we conducted a pilot study to generate and qualitatively evaluate feedback for a single question.
This allowed us to determine the final set of elements based on their distinct educational characteristics, with each adding a specific emphasis (e.g., instructional specificity or information coverage) to the baseline Normal condition.

We use gpt-5-2025-08-07 (GPT-5)~\cite{GPT-5} to generate all feedback in advance, and produce indirect feedback that encourages learners to derive the correct answer without explicitly revealing it.
See Appendix~\ref{app:feedback_element} for details.

\paragraph{Normal}
We use this design as the reference condition for comparison across feedback elements.
This explicitly indicates whether the learner's response is correct or incorrect and provides a standard explanation aimed at bridging gaps in the learner's knowledge and reasoning. 

\paragraph{Keywords}
This design focuses on information granularity for learner understanding.
Rather than presenting keywords alone, the feedback highlights particularly important terms within the explanation using brackets.

\paragraph{Actionability}
This design investigates the impact of instructional specificity on answer revision.
Since abstract guidance may make it difficult for learners to judge how to improve their responses, the feedback provides concrete revision steps and checklists that indicate what should be reviewed.

\paragraph{Novelty}
This design explores introducing information beyond the learned content.
Regardless of the answer's correctness, the feedback provides advanced perspectives.
\begin{figure*}[t]
\includegraphics[width=\textwidth]{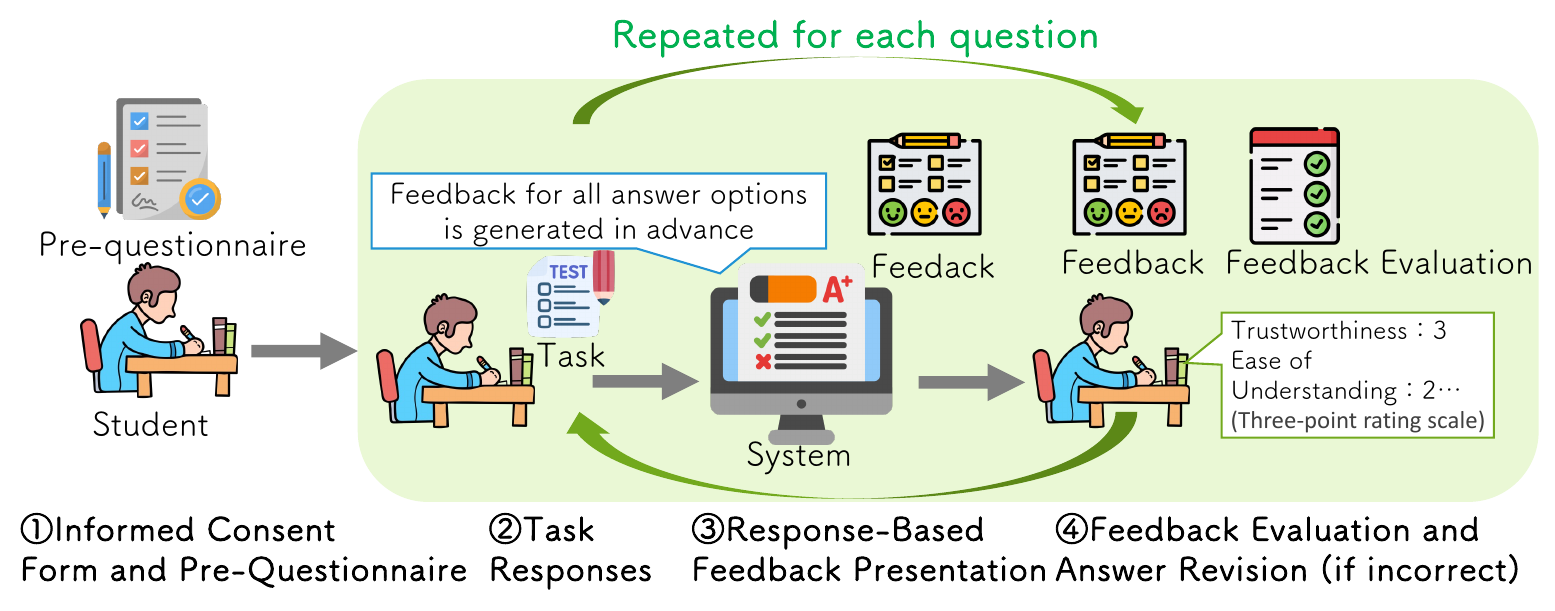}
\caption{
Overview of the experimental procedure.
Participants first provide informed consent and complete a pre questionnaire.
They then answer the task questions and evaluate feedback using a three-point scale across six criteria, such as trustworthiness and ease of understanding.
If they make mistakes in their responses, they repeat the cycle of answering, reviewing the feedback, and evaluating it until they reach the correct answer.
We randomly assign a feedback element to each question and keep it fixed throughout the revision process.
} 
\label{fig:overflow}
\end{figure*}

\paragraph{Coverage}
This design examines how informational completeness influences learner understanding.
Rather than focusing on a single aspect, the feedback describes all logical elements required to reach the correct answer.
\paragraph{Positivity}
This design investigates the influence of affective factors on learning.
In addition to pointing out errors, the feedback incorporates positive expressions, such as praise for partially correct reasoning and encouraging comments.

\subsection{Qualitative Analysis of Feedback}

\label{ssec:qualitative_analysis}
We generate a total of 222 feedback instances for all 37 answer options.
To characterize the educational elements of the generated feedback, we qualitatively analyze all instances along five dimensions: Specificity, Information Density, Cognitive Scaffolding, Interaction Framing, and Educational Reliability.
The analysis suggests potential trade-offs among affective support, informational detail, and inference preservation.

\textit{Positivity} is dominated by praise-oriented expressions, with more than half of its instances lacking concrete corrective guidance about what learners should revise.
In contrast, \textit{Novelty} frequently provides detailed and knowledge-rich explanations, yet nearly 40\% (14 out of 37) of its instances contain limited task-specific revision guidance because explanations often extend beyond the scope of the target question.
Regarding Educational Reliability, approximately 13\% (29 out of 222) of all feedback instances indirectly suggest the correct answer despite not explicitly revealing it.
Notably, about 60\% of these cases (18 out of 29) involve \textit{Actionability} or \textit{Coverage} (9 cases each), where detailed explanations often narrow the space of possible answers.
This tendency appears particularly often in correspondence-based questions.
This suggests that answer leakage may depend not only on feedback elements but also on question format.

Because such answer leakage may affect the validity of revision performance analysis, we exclude these instances from subsequent quantitative analyses.
As a result, 88\% of first-attempt responses and 83\% of feedback--response instances remain for analysis.
See Appendix~\ref{sec:feedback_example} for feedback examples, Appendix~\ref{app:qualitative_analysis} for details of the qualitative analysis and representative examples, and Appendix~\ref{app:sensitivity_leaked} for the rationale for excluding answer-leaking feedback and the corresponding sensitivity analysis.

\section{Empirical Experiment}
\label{sec:experiment}
\subsection{Participants}
\label{ssec:participants}
Nine high school basic biology classes participated in this study. 
The participants were approximately 15--16 years old and enrolled in academically selective high schools. 
We assigned this activity as winter-break homework.
Students were informed that declining consent to data use would have no academic consequences. 
Informed consent was obtained from all participants, and parental consent was also obtained prior to the commencement of the experiment. We obtained data from 321 students who provided consent for the use of their learning logs. This study was approved by the ethics committees of our university and research institute. See Appendix~\ref{app:participants} and~\ref{app:ethic} for details.

\subsection{Dataset}
\label{ssec:dataset}
We use the biology subset of the dataset released by~\citet{2025.EDM.poster-demo-papers.281}, which is derived from Japan's National Center Test for University Admissions.
These questions are designed for university entrance examinations and have an average correct response rate of approximately 55--70\% among third-year high school students.
The dataset consists of multiple-choice questions, in which students select one or more correct answers.
We exclude multiple-answer questions due to the large number of possible answer combinations.
Since participants have not completed all biology units, we select questions aligned with the basic biology curriculum through a two-stage filtering process involving LLMs and expert review.
This results in seven questions selected from the original 101 questions.
See Appendix~\ref{app:dataset} for filtering process.

\subsection{Learning Data Collection}
\label{ssec:data_collection}
Figure~\ref{fig:overflow} shows the overall experimental procedure.
First, we explain the purpose of the study and the handling of collected data.
After obtaining informed consent, participants complete a pre-questionnaire administered via Google Forms.
The questionnaire includes a Japanese Big Five personality scale~\cite{Murakami1997BigFive}, along with items assessing participants' experience with LLMs and background in basic biology.
Next, participants access a web-based system developed by the authors using their assigned student IDs and passwords.
After logging in, participants view a list of tasks and can select questions to answer.
The order in which the questions are presented is randomized for each participant.
They then submit their responses and immediately receive feedback corresponding to their selected option.

\paragraph{Subjective Evaluation Criteria}
After reviewing the feedback, participants evaluate it on a three-point scale across six criteria: the perceived reliability of feedback (Trustworthiness),
its usefulness as a guide for answer revision (Review Guideline),
the quality of expression (Expression Quality),
ease of comprehension (Ease of Understanding),
clarity of questions' key points (Key Points Clarity),
and the extent to which it provides new or additional knowledge (New Knowledge).

We select six representative criteria based on characteristics of effective educational feedback consistently identified in prior educational research~\cite{black1998assessment,Hattie2007,Valerie2008,lipnevich2016toward,carless2018development,wisniewski2020power}, while also considering participant workload.
The criteria are designed to correspond to the educational feedback elements examined in our study, enabling us to assess the effects of different feedback designs from multiple perspectives.

Participants iteratively revise their responses and review and evaluate the feedback until they reach the correct answer.
Throughout this process, we collect all responses, feedback histories, and evaluation results as learning logs.
Appendix~\ref{app:UI} illustrates the system interface and evaluation criteria.

\begin{figure}[t]
\includegraphics[width=\linewidth]{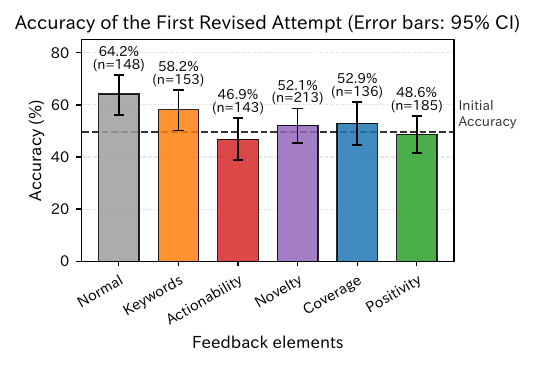}
\caption{
Accuracy of the first revised attempt.
\textit{Normal}, \textit{Keywords}, and \textit{Coverage} are the most effective methods for supporting answer revision, in this order.
}
\label{fig:acc}
\end{figure}

\begin{figure*}[t]
\includegraphics[width=\textwidth]{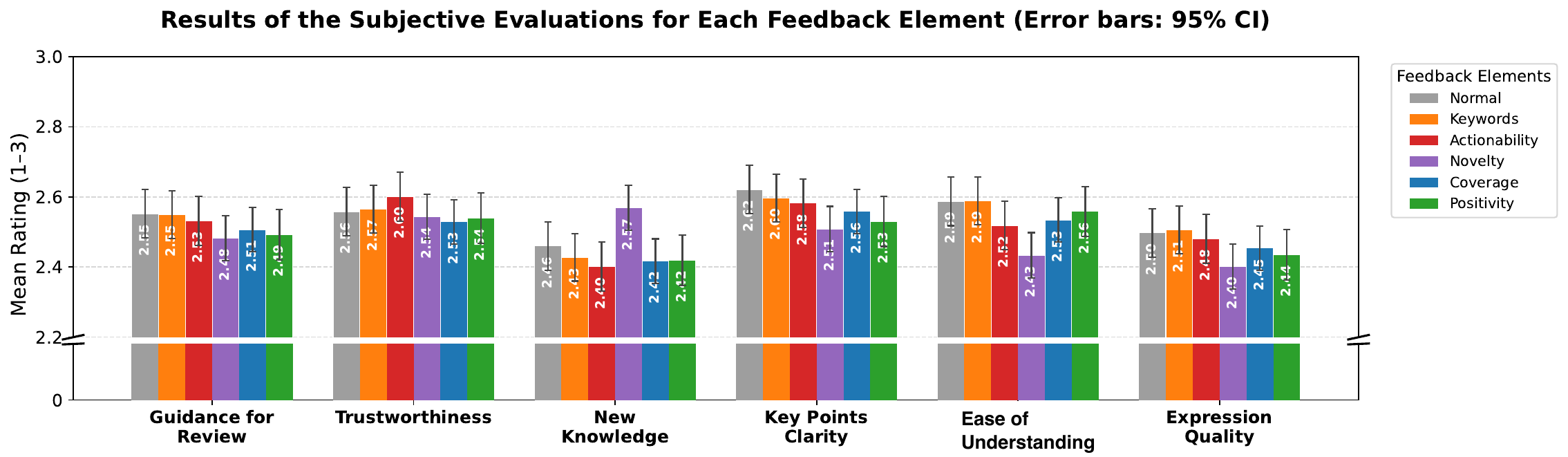}
\caption{
Results of the subjective evaluations for each feedback element.
\textit{Keywords}, \textit{Normal}, and \textit{Actionability} tend to receive higher ratings across many criteria.
} 
\label{fig:student_fb_evaluation_results}
\end{figure*}
\begin{table*}[t]
\centering
\setlength{\tabcolsep}{3pt}
\resizebox{\linewidth}{!}{
\begin{tabular}{lllllll}
\toprule
\multicolumn{1}{c}{\textbf{Elements}}
& \multicolumn{1}{c}{\textbf{Guidance}}
& \multicolumn{1}{c}{\textbf{Trust}}
& \multicolumn{1}{c}{\textbf{Knowledge}}
& \multicolumn{1}{c}{\textbf{Key Points}}
& \multicolumn{1}{c}{\textbf{Understanding}}
& \multicolumn{1}{c}{\textbf{Expression}} \\
\midrule

Intercept (Normal.)
& 1.40 [0.92, 2.13]
& 1.76 [1.13, 2.75]$^{*}$
& 1.42 [0.98, 2.05]
& 1.65 [1.09, 2.51]$^{*}$
& 1.73 [1.16, 2.60]$^{**}$
& 1.40 [0.92, 2.12] \\

\addlinespace

\rowcolor{LightGray} Keywords
& 1.18 [0.76, 1.82]
& 0.75 [0.52, 1.09]
& 0.76 [0.50, 1.14]
& 1.00 [0.66, 1.51]
& 0.77 [0.52, 1.15]
& 0.76 [0.51, 1.14] \\

Actionability
& 1.07 [0.67, 1.71]
& 0.72 [0.48, 1.08]
& 0.86 [0.57, 1.30]
& 0.97 [0.62, 1.51]
& 0.80 [0.53, 1.20]
& 0.83 [0.54, 1.28] \\

\rowcolor{LightGray} Novelty
& 0.62 [0.40, 0.95]$^{*}$
& 0.55 [0.34, 0.87]$^{*}$
& 0.74 [0.48, 1.13]
& 0.60 [0.40, 0.90]$^{*}$
& 0.47 [0.30, 0.73]$^{***}$
& 0.46 [0.29, 0.74]$^{**}$ \\

Coverage
& 1.20 [0.80, 1.79]
& 0.80 [0.53, 1.20]
& 0.84 [0.54, 1.30]
& 1.08 [0.74, 1.58]
& 0.96 [0.63, 1.46]
& 0.84 [0.53, 1.33] \\

\rowcolor{LightGray} Positivity
& 0.86 [0.56, 1.31]
& 0.58 [0.41, 0.83]$^{**}$
& 0.75 [0.51, 1.10]
& 0.75 [0.49, 1.13]
& 0.72 [0.48, 1.09]
& 0.71 [0.47, 1.06] \\

\midrule

Wald $\chi^2$
& 14.967
& 12.380
& 3.249
& 16.426
& 18.598
& 14.231 \\

Wald $p$
& 0.0105
& 0.0299
& 0.6617
& 0.0057
& 0.0023
& 0.0142 \\

\bottomrule
\end{tabular}
}
\vspace{-0.5mm}
\caption{Logistic GEE results for subjective evaluations across six criteria. Odds ratios (ORs) and 95\% confidence intervals are reported relative to the \textit{Normal} condition. $^{*}p < .05$, $^{**}p < .01$, $^{***}p < .001$.}
\label{tb:logic_gee_subjective_results}
\end{table*}

\section{Results}
\label{sec:results}
We collect 2,197 first attempts and 4,478 feedback--response instances.
As described in §~\ref{ssec:qualitative_analysis}, we exclude responses associated with answer-leaking feedback from the analysis.
Using the remaining data, we analyze the effects of feedback elements on immediate revision performance and subjective evaluations for the full participant sample. 
We use immediate revision performance as a proxy for short-term support rather than durable learning gains. 
Finally, we examine how these effects vary across learner profiles. 

\subsection{Answer Revision Results}
\label{ssec:answer_revision}
To investigate~\textit{RQ1: Which feedback elements most effectively improve revision performance?}, we analyze the effects of feedback elements on immediate revision performance.
The mean accuracy of learners' initial responses is 49.6\%, which indicates that the tasks are of appropriate difficulty.

\paragraph{First Revise Accuracy with Feedback}
We aim to identify feedback elements that lead learners to correct answers.
To this end, we focus on participants whose first response is incorrect and compute the accuracy of their revised responses after receiving feedback.
Figure~\ref{fig:acc} shows the accuracy of the first revised attempt for each element.
This accuracy ranges from 46.9\% to
64.2\%.
\textit{Normal} shows the highest first-revision accuracy, followed by \textit{Keywords} and \textit{Coverage}.
We first apply a chi-square test to examine whether accuracy differs across feedback elements.
The results reveal a significant association between the element and accuracy ($\chi^2 (5)=12.611, p=.027$).
To account for repeated observations from the same learner across different questions, we further conduct a Generalized Estimating Equations (GEE) analysis, which models within-learner correlations and provides more reliable estimates.
Using \textit{Normal} as the reference, the results indicate that \textit{Actionability}, \textit{Novelty}, and \textit{Positivity} are associated with significantly lower accuracy rates ($p<.05$).
\textit{Coverage} shows a marginally significant negative association ($p=.061$), whereas \textit{Keywords} does not differ significantly from \textit{Normal}.
See Appendix~\ref{app:gee}.




\subsection{Overall Feedback Evaluation Results}
\label{ssec:overall_feedback_evaluation}
To investigate \textit{RQ2: Which feedback elements lead to more favorable subjective evaluations?}, we analyze subjective evaluations.
Figure~\ref{fig:student_fb_evaluation_results} shows the subjective evaluation results for each element.
We examine which feedback elements receive higher ratings.
\textit{Normal} ranks first in three of the six criteria and appears in the top three for all criteria.
\textit{Keywords} and \textit{Actionability} also rank within the top three for all and four criteria, respectively.

To examine statistical significance, we apply logistic GEE, as ratings from the same learner across different questions and feedback elements violate the independence assumption of standard regression.
Because the subjective evaluations use a three-point scale with limited variance, we dichotomize the ratings into Agree (3) and Not Agree (1--2) before fitting the model.
We first attempt an ordinal GEE, but it fails to converge because of quasi-complete separation caused by the strong skew toward Agree responses.
We report odds ratios (ORs), obtained by exponentiating the logistic GEE coefficients, together with 95\% confidence intervals, using \textit{Normal} as the reference condition.

As shown in Table~\ref{tb:logic_gee_subjective_results}, the main directional patterns observed in the original analysis are preserved.
In particular, \textit{Novelty} shows significantly lower odds of receiving an Agree rating than \textit{Normal} for five of the six criteria (OR = 0.46--0.62), while \textit{Positivity} shows significantly lower odds only for Trustworthiness (OR = 0.58).
See Appendix~\ref{app:stat_subject}.

Overall, \textit{Normal} and \textit{Keywords} show relatively favorable performance across both revision accuracy and subjective evaluations.
\textit{Coverage} shows relatively high revision accuracy, whereas \textit{Actionability} tends to receive favorable subjective evaluations.
In contrast, \textit{Novelty} receives less favorable subjective evaluations across most criteria, while the negative effect of \textit{Positivity} is more limited.

\subsection{Feedback Evaluation by Big Five Traits}

To investigate \textit{RQ3: How do subjective evaluations of feedback elements vary across learners with different Big Five personality traits?}, we examine how learners' subjective feedback evaluations of feedback elements vary by their Big Five traits.

\begin{figure}[t]
\includegraphics[width=\linewidth]{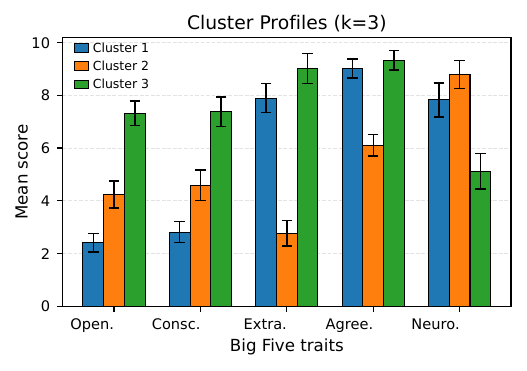}
\caption{Profiles of three learner clusters based on Big Five traits.
Cluster~1 shows high Agreeableness and Extraversion, along with low Openness and Conscientiousness.
Cluster~2 is characterized by higher Neuroticism and lower Extraversion than the other clusters.
Cluster~3 exhibits high levels across all traits except Neuroticism.}
\label{fig:bar_graph_k3}
\end{figure}

\begin{figure*}[t]
\includegraphics[width=\textwidth]{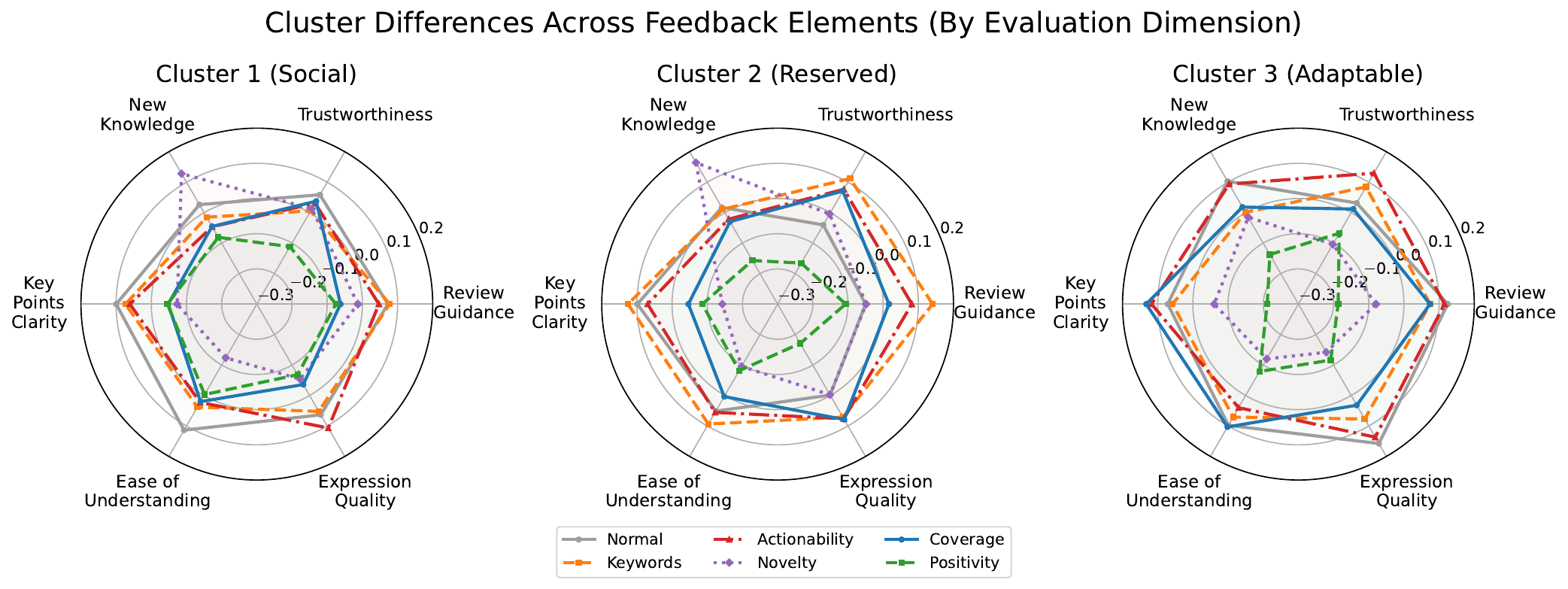}
\caption{Feedback evaluation results by cluster. Values are standardized for each learner and each evaluation dimension. In \texttt{Social}, ``Trustworthiness'' scores are similar across all feedback elements. In \texttt{Reserved}, \textit{Keywords} receives high ratings, while in \texttt{Adaptable}, \textit{Keywords}, \textit{Normal}, and \textit{Actionability} receive high ratings.}

\label{fig:radar_grid_p1}
\end{figure*}

\subsubsection{Learner Personality Clustering}
\label{sssec:clustering}

We examine how learners with different Big Five personality traits evaluate feedback elements.
Because analyzing each trait individually is complex, we instead identify common patterns by grouping learners into representative personality profiles.
First, we compute the scale score for each trait from a questionnaire~\cite{Murakami1997BigFive}.
We represent each learner as a five-dimensional personality vector and apply k-means clustering to group learners with similar profiles.
The elbow result indicates that three or four clusters are appropriate.
Based on interpretability and consistency with previous studies~\cite{asendorpf2002puzzle,wall2019personality,takami2023personality}, we adopt three clusters.
Figure~\ref{fig:bar_graph_k3} shows the three learner cluster profiles.  
Cluster~1 shows relatively high extraversion and agreeableness and lower levels of openness and conscientiousness.  
Cluster~2 is characterized by notably low extraversion and relatively high neuroticism.  
Cluster~3 shows high levels of openness, conscientiousness, and agreeableness.
Based on these profiles, we label Cluster~1 as \texttt{Social}, Cluster~2 as \texttt{Reserved}, and Cluster~3 as \texttt{Adaptable}.
See Appendix~\ref{app:elbow} for details.

\subsubsection{Feedback Evaluation by Cluster}
\label{sssec:evaluation_result}

Figure~\ref{fig:radar_grid_p1} shows standardized subjective evaluation scores for each feedback element and evaluation criterion.
Several elements exhibit relatively consistent evaluation patterns across clusters.
\textit{Keywords} generally receives favorable evaluations across all clusters, whereas  \textit{Positivity} tends to receive relatively low evaluations.
These patterns suggest that task-focused feedback may be broadly well received, while praise-oriented feedback alone may be less favorably perceived in the present setting.

We also find several cluster-specific tendencies.
Both \texttt{Social} and \texttt{Reserved} show relatively high New Knowledge ratings for \textit{Novelty}, whereas this tendency is less pronounced in \texttt{Adaptable}.
In \texttt{Social}, Trustworthiness ratings are relatively similar across feedback elements, and \textit{Normal}, \textit{Keywords}, and \textit{Actionability} receive relatively balanced evaluations across criteria.
In \texttt{Reserved}, \textit{Keywords} receives consistently high ratings, while \textit{Positivity} and \textit{Novelty} receive relatively low ratings for Ease of Understanding and Key Points Clarity.
In \texttt{Adaptable}, relatively favorable evaluations are distributed across a broader range of elements, including \textit{Normal}, \textit{Keywords}, \textit{Actionability}, and \textit{Coverage}.
These results suggest that possible heterogeneity in how learners perceive different feedback designs rather than establishing distinct element preferences by personality profile.

To examine whether overall subjective evaluation patterns differ across clusters, we conduct Kruskal--Wallis tests using cluster as the independent factor for each evaluation criterion.
The results show significant differences across clusters for all six criteria (\textit{p} < .05). 
We then perform Dunn's post hoc tests with Bonferroni correction to identify specific pairwise differences.
For Trustworthiness, significant differences appear between \texttt{Reserved} and both \texttt{Social} and \texttt{Adaptable} (both \textit{p} < .01).
Significant pairwise differences between \texttt{Reserved} and the other clusters are also observed for New Knowledge and Expression Quality.
For Review Guidance, although the overall Kruskal--Wallis test is significant, no pairwise differences remain significant after Bonferroni correction.

Overall, subjective evaluation patterns differ across learner clusters, suggesting heterogeneity in how feedback is perceived. 
However, these patterns should not be taken as evidence that personality profiles systematically prefer feedback elements.


\section{Analysis}
\label{sec:analysis}

We investigate how feedback elements affect both immediate revision performance and subjective evaluations, followed by an analysis of how these effects vary across learner profiles.

\subsection{Feedback Element Effects}
\label{ssec:fb_component}

\paragraph{Immediate Revision Performance}
\textit{Normal} leads to the highest performance, followed by \textit{Keywords} and \textit{Coverage}.
\textit{Normal} clearly indicates both what needs to be corrected and how to revise, while \textit{Coverage} provides comprehensive information but spreads learners' attention across multiple points.

The higher performance of the \textit{Normal} condition, particularly in the first revision attempt, may be attributed to the directness and conciseness of the feedback.
By explicitly identifying learners' errors and providing the information necessary for correction, \textit{Normal} may have been well suited to immediate answer revision.
In contrast, additional feedback elements provide richer educational information but also increase the amount of information that learners need to process.
Under the setting examined in this study, in which learners revise their answers immediately after receiving feedback, such information may therefore have produced less direct effects on revision performance.

Importantly, this result does not imply that feedback containing additional information is educationally less useful.
Feedback that is effective for immediate error correction may differ from feedback that supports deeper understanding or longer-term knowledge retention.
Thus, the relative effectiveness of different feedback designs may depend on the learning objective and the timing of evaluation, and this possibility should be further investigated in future work.

\paragraph{Subjective Evaluation Performance}
\textit{Normal}, \textit{Keywords}, and \textit{Actionability} receive relatively high subjective ratings. However, \textit{Actionability}'s lower revision accuracy suggests that learners may favor feedback that feels supportive and actionable even when it does not immediately help them reach the correct answer. This divergence suggests that perceived usefulness and immediate revision success capture different aspects of feedback effectiveness. In contrast, \textit{Novelty} receives less favorable ratings across several criteria, while \textit{Positivity} receives a significantly lower rating for Trustworthiness. These patterns are consistent with evidence that feedback effectiveness depends on its informational content and that praise-oriented feedback is not necessarily beneficial~\cite{Hattie2007,wisniewski2020power}.

\subsection{Personality-aware Feedback Evaluation}
\label{ssec:Personality_aware}

We observe differences in feedback evaluation patterns across Big Five-based learner clusters.
\texttt{Social} learners (high extraversion/agreeableness) tend to rate validity- and trust-related aspect relatively highly, while \textit{Normal} and \textit{Keywords} receive comparatively favorable evaluations.
This pattern may indicate that clear rationales and revision-oriented guidance are valued within this cluster.
\texttt{Reserved} learners (low extraversion/ high neuroticism) tend to rate \textit{Positivity} and \textit{Novelty} less favorably, while \textit{Keywords} and \textit{Normal} receive relatively high evaluations.
This pattern may be related to prior findings linking neuroticism to avoidance-oriented academic motivation~\cite{komarraju2005relationship}.
\texttt{Adaptable} learners (high openness/conscientiousness/agreeableness) show relatively favorable evaluations across a broader range of feedback elements, including \textit{Actionability}.
This tendency may be consistent with prior work linking these traits to intrinsic academic motivation and  achievement~\cite{komarraju2009role}.
Overall, feedback evaluations are not uniform across learner clusters, although the observed element-specific differences should be interpreted cautiously.

\subsection{Implications for LLM Feedback}
\label{ssec:LLM-Based_Educational_Systems}

\paragraph{Learner Simulation and Feedback Design}
Recent studies have explored using LLMs to simulate human preferences and behaviors, including educational settings~\cite{liu-etal-2024-personality,sanyal-etal-2025-investigating,hu2026simbench}. Prior studies on automatic feedback evaluation often rely on LLM-based evaluations~\cite{qian2025deanllmtutorsexploring,chu2026feedevalpedagogicallyalignedevaluation}. 
We empirically investigate how learners perceive and respond to different feedback elements and our findings may support future research on whether LLMs can simulate learner preferences and revision behaviors more realistically. 
Such simulations could enable scalable evaluation of feedback strategies before large-scale human experiments. 
This suggests the potential of empirically grounded learner simulation for personalized feedback design.

\paragraph{Responsible Use of LLM Feedback}
While LLM-generated feedback may support personalized learning, its broader use in educational settings also requires careful consideration.
Excessive reliance on AI may reduce learners' opportunities for independent engagement, and the long-term effects of such use remain unclear~\cite{beale2025revolutionarrivedcurrentstate}.
Controlled empirical studies are therefore important for identifying when and how generative AI can support learning without undermining learners' own problem-solving processes.
Future work should therefore consider not only the content of AI-generated feedback, but also when and how it is integrated into the learning process.

The use of learner data also raises privacy concerns, particularly when psychological or personality-related information is collected from minors.
Drawing on privacy-by-design frameworks for child-facing LLMs~\cite{addae2026privacydesignframeworklarge}, future work should consider safeguards such as data minimization, access control, purpose limitation, and age-appropriate consent.

\section{Conclusion}
\label{sec:conclusion}
We define six feedback elements for high school biology multiple-choice questions and evaluate their effects on immediate revision performance and subjective evaluations through an empirical experiment with 321 students.
Our results show that feedback presenting task-relevant information clearly is associated with better immediate revision.
Learners also tend to evaluate feedback more favorably when it is practical, understandable, and easy to apply.
We further observe differences in subjective evaluation patterns across learner profiles, particularly for informational novelty and affective framing.
These findings suggest that learner characteristics may be associated with differences in feedback evaluation, motivating further investigation of learner-aware personalization.


\section*{Limitations}
Our study has several limitations.
\paragraph{Task and Question Generalizability}
We focus exclusively on biology tasks with a limited number of questions, and the generalizability of our findings to other subject domains remains unclear.
Although the questions were reviewed by education experts for their appropriateness, question-level characteristics such as difficulty were not systematically controlled.
In addition, the current feedback design does not account for question-level accuracy or difficulty. Future work should explicitly incorporate these characteristics during question selection and feedback design, which may enable more adaptive feedback tailored to K--12 learners.

\paragraph{Learner Characteristics}
Our analysis focuses on personality traits and does not account for other learner characteristics, such as prior knowledge, academic ability, or learning history, which may also influence feedback preferences and effectiveness. 
Future work should incorporate these additional factors to provide a more comprehensive understanding of how learner characteristics shape responses to different feedback designs.

\paragraph{Model Generalizability}
We use only GPT-5 to generate feedback because it consistently produced the highest-quality feedback in our pilot evaluation compared to other LLMs.
Nevertheless, it is unclear whether feedback generated by other LLMs would yield similar effects.

\paragraph{Effectiveness Measures}
Our effectiveness measures---correctness of revised answers and subjective evaluations---may favor feedback that more explicitly guides learners toward correct answers, rather than reflecting deeper learning gains.
Feedback designs that are effective for immediate answer revision may differ from those that support long-term understanding and knowledge retention.
Future work should therefore consider longer-term learning outcomes as a more robust measure of feedback effectiveness.

\paragraph{Control of External Assistance}
External assistance was not explicitly controlled during the experiment. 
According to the supervising teacher, students were informed that the activity was intended to review previously learned biology content and that their performance would not affect their grades, which likely reduced their incentive to seek external help.
In addition, the experiment was conducted during the winter break making peer discussion less likely.
Nevertheless, we cannot rule out the possibility that students used external resources or received assistance from others.
Future studies should therefore explicitly control or record external assistance, ideally under conditions that more closely resemble an actual classroom setting.

\paragraph{Responsible Use of Generative AI}
Our study does not fully address broader risks associated with the use of generative AI in education. These include privacy concerns, particularly when collecting psychological data from minors, and the risk that excessive reliance on AI may reduce learners' independent engagement. Future work should therefore examine not only feedback content, but also when and how generative AI should be used in learning.

\section*{Ethical Considerations}
Our study involves empirical experiments with first-year high school students. 
The activity was assigned as mandatory winter-break homework; however, consent to the use of collected data was entirely voluntary and independent of the homework requirement. Students were explicitly informed that declining consent to data use would have no academic consequences. Informed consent was obtained from all participants, and parental consent was also obtained prior to the commencement of the experiment. 
Data were collected using pseudonymous student IDs and anonymized before analysis and used solely for research purposes. This study was approved by the institutional ethics review board of the authors' institution.
We used GPT-5~\cite{GPT-5} solely for feedback generation, and no personally identifiable information was included in any prompts submitted to the model.
See Appendix~\ref{app:ethic}.

\paragraph{Use of AI Assistants}
We used generative AI tools to assist with code debugging and modification, literature survey support, and English writing revision. 
The AI tools were used only as supportive assistants for improving readability and development efficiency, and all experimental designs, analyses, interpretations, and final manuscript contents were verified and finalized by the authors.

\section*{Acknowledgments}
The authors would like to thank the anonymous
reviewers for their helpful comments. 
This work was supported by 
JST FOREST Grant Number JPMJFR232R, 
JST BOOST Grant Number JPMJBS2421, JPMJBY24D9 and 
JSPS KAKENHI Grant Numbers JP23K17012, JP23K25698, and partially funded by the Cabinet Office, Government of Japan, under the BRIDGE initiative through the MEXT measure ``Building Educational AI Models and Developing a Human-Agent Collaborative AI Ecosystem for Educational Support.''
In this work, we used the ``mdx: a platform for building data-empowered society.''

\bibliography{references}

@misc{GPT-5,
      title={{OpenAI GPT-5 System Card}}, 
      author={Aaditya Singh and Adam Fry and Adam Perelman and Adam Tart and Adi Ganesh and Ahmed El-Kishky and Aidan McLaughlin and Aiden Low and AJ Ostrow and Akhila Ananthram and Akshay Nathan and Alan Luo and Alec Helyar and Aleksander Madry and Aleksandr Efremov and Aleksandra Spyra and Alex Baker-Whitcomb and Alex Beutel and Alex Karpenko and Alex Makelov and Alex Neitz and Alex Wei and Alexandra Barr and Alexandre Kirchmeyer and Alexey Ivanov and Alexi Christakis and Alistair Gillespie and Allison Tam and Ally Bennett and Alvin Wan and Alyssa Huang and Amy McDonald Sandjideh and Amy Yang and Ananya Kumar and Andre Saraiva and Andrea Vallone and Andrei Gheorghe and Andres Garcia Garcia and Andrew Braunstein and Andrew Liu and Andrew Schmidt and Andrey Mereskin and Andrey Mishchenko and Andy Applebaum and Andy Rogerson and Ann Rajan and Annie Wei and Anoop Kotha and Anubha Srivastava and Anushree Agrawal and Arun Vijayvergiya and Ashley Tyra and Ashvin Nair and Avi Nayak and Ben Eggers and Bessie Ji and Beth Hoover and Bill Chen and Blair Chen and Boaz Barak and Borys Minaiev and Botao Hao and Bowen Baker and Brad Lightcap and Brandon McKinzie and Brandon Wang and Brendan Quinn and Brian Fioca and Brian Hsu and Brian Yang and Brian Yu and Brian Zhang and Brittany Brenner and Callie Riggins Zetino and Cameron Raymond and Camillo Lugaresi and Carolina Paz and Cary Hudson and Cedric Whitney and Chak Li and Charles Chen and Charlotte Cole and Chelsea Voss and Chen Ding and Chen Shen and Chengdu Huang and Chris Colby and Chris Hallacy and Chris Koch and Chris Lu and Christina Kaplan and Christina Kim and CJ Minott-Henriques and Cliff Frey and Cody Yu and Coley Czarnecki and Colin Reid and Colin Wei and Cory Decareaux and Cristina Scheau and Cyril Zhang and Cyrus Forbes and Da Tang and Dakota Goldberg and Dan Roberts and Dana Palmie and Daniel Kappler and Daniel Levine and Daniel Wright and Dave Leo and David Lin and David Robinson and Declan Grabb and Derek Chen and Derek Lim and Derek Salama and Dibya Bhattacharjee and Dimitris Tsipras and Dinghua Li and Dingli Yu and DJ Strouse and Drew Williams and Dylan Hunn and Ed Bayes and Edwin Arbus and Ekin Akyurek and Elaine Ya Le and Elana Widmann and Eli Yani and Elizabeth Proehl and Enis Sert and Enoch Cheung and Eri Schwartz and Eric Han and Eric Jiang and Eric Mitchell and Eric Sigler and Eric Wallace and Erik Ritter and Erin Kavanaugh and Evan Mays and Evgenii Nikishin and Fangyuan Li and Felipe Petroski Such and Filipe de Avila Belbute Peres and Filippo Raso and Florent Bekerman and Foivos Tsimpourlas and Fotis Chantzis and Francis Song and Francis Zhang and Gaby Raila and Garrett McGrath and Gary Briggs and Gary Yang and Giambattista Parascandolo and Gildas Chabot and Grace Kim and Grace Zhao and Gregory Valiant and Guillaume Leclerc and Hadi Salman and Hanson Wang and Hao Sheng and Haoming Jiang and Haoyu Wang and Haozhun Jin and Harshit Sikchi and Heather Schmidt and Henry Aspegren and Honglin Chen and Huida Qiu and Hunter Lightman and Ian Covert and Ian Kivlichan and Ian Silber and Ian Sohl and Ibrahim Hammoud and Ignasi Clavera and Ikai Lan and Ilge Akkaya and Ilya Kostrikov and Irina Kofman and Isak Etinger and Ishaan Singal and Jackie Hehir and Jacob Huh and Jacqueline Pan and Jake Wilczynski and Jakub Pachocki and James Lee and James Quinn and Jamie Kiros and Janvi Kalra and Jasmyn Samaroo and Jason Wang and Jason Wolfe and Jay Chen and Jay Wang and Jean Harb and Jeffrey Han and Jeffrey Wang and Jennifer Zhao and Jeremy Chen and Jerene Yang and Jerry Tworek and Jesse Chand and Jessica Landon and Jessica Liang and Ji Lin and Jiancheng Liu and Jianfeng Wang and Jie Tang and Jihan Yin and Joanne Jang and Joel Morris and Joey Flynn and Johannes Ferstad and Johannes Heidecke and John Fishbein and John Hallman and Jonah Grant and Jonathan Chien and Jonathan Gordon and Jongsoo Park and Jordan Liss and Jos Kraaijeveld and Joseph Guay and Joseph Mo and Josh Lawson and Josh McGrath and Joshua Vendrow and Joy Jiao and Julian Lee and Julie Steele and Julie Wang and Junhua Mao and Kai Chen and Kai Hayashi and Kai Xiao and Kamyar Salahi and Kan Wu and Karan Sekhri and Karan Sharma and Karan Singhal and Karen Li and Kenny Nguyen and Keren Gu-Lemberg and Kevin King and Kevin Liu and Kevin Stone and Kevin Yu and Kristen Ying and Kristian Georgiev and Kristie Lim and Kushal Tirumala and Kyle Miller and Lama Ahmad and Larry Lv and Laura Clare and Laurance Fauconnet and Lauren Itow and Lauren Yang and Laurentia Romaniuk and Leah Anise and Lee Byron and Leher Pathak and Leon Maksin and Leyan Lo and Leyton Ho and Li Jing and Liang Wu and Liang Xiong and Lien Mamitsuka and Lin Yang and Lindsay McCallum and Lindsey Held and Liz Bourgeois and Logan Engstrom and Lorenz Kuhn and Louis Feuvrier and Lu Zhang and Lucas Switzer and Lukas Kondraciuk and Lukasz Kaiser and Manas Joglekar and Mandeep Singh and Mandip Shah and Manuka Stratta and Marcus Williams and Mark Chen and Mark Sun and Marselus Cayton and Martin Li and Marvin Zhang and Marwan Aljubeh and Matt Nichols and Matthew Haines and Max Schwarzer and Mayank Gupta and Meghan Shah and Melody Y. Guan and Melody Huang and Meng Dong and Mengqing Wang and Mia Glaese and Micah Carroll and Michael Lampe and Michael Malek and Michael Sharman and Michael Zhang and Michele Wang and Michelle Pokrass and Mihai Florian and Mikhail Pavlov and Miles Wang and Ming Chen and Mingxuan Wang and Minnia Feng and Mo Bavarian and Molly Lin and Moose Abdool and Mostafa Rohaninejad and Nacho Soto and Natalie Staudacher and Natan LaFontaine and Nathan Marwell and Nelson Liu and Nick Preston and Nick Turley and Nicklas Ansman and Nicole Blades and Nikil Pancha and Nikita Mikhaylin and Niko Felix and Nikunj Handa and Nishant Rai and Nitish Keskar and Noam Brown and Ofir Nachum and Oleg Boiko and Oleg Murk and Olivia Watkins and Oona Gleeson and Pamela Mishkin and Patryk Lesiewicz and Paul Baltescu and Pavel Belov and Peter Zhokhov and Philip Pronin and Phillip Guo and Phoebe Thacker and Qi Liu and Qiming Yuan and Qinghua Liu and Rachel Dias and Rachel Puckett and Rahul Arora and Ravi Teja Mullapudi and Raz Gaon and Reah Miyara and Rennie Song and Rishabh Aggarwal and RJ Marsan and Robel Yemiru and Robert Xiong and Rohan Kshirsagar and Rohan Nuttall and Roman Tsiupa and Ronen Eldan and Rose Wang and Roshan James and Roy Ziv and Rui Shu and Ruslan Nigmatullin and Saachi Jain and Saam Talaie and Sam Altman and Sam Arnesen and Sam Toizer and Sam Toyer and Samuel Miserendino and Sandhini Agarwal and Sarah Yoo and Savannah Heon and Scott Ethersmith and Sean Grove and Sean Taylor and Sebastien Bubeck and Sever Banesiu and Shaokyi Amdo and Shengjia Zhao and Sherwin Wu and Shibani Santurkar and Shiyu Zhao and Shraman Ray Chaudhuri and Shreyas Krishnaswamy and Shuaiqi and Xia and Shuyang Cheng and Shyamal Anadkat and Simón Posada Fishman and Simon Tobin and Siyuan Fu and Somay Jain and Song Mei and Sonya Egoian and Spencer Kim and Spug Golden and SQ Mah and Steph Lin and Stephen Imm and Steve Sharpe and Steve Yadlowsky and Sulman Choudhry and Sungwon Eum and Suvansh Sanjeev and Tabarak Khan and Tal Stramer and Tao Wang and Tao Xin and Tarun Gogineni and Taya Christianson and Ted Sanders and Tejal Patwardhan and Thomas Degry and Thomas Shadwell and Tianfu Fu and Tianshi Gao and Timur Garipov and Tina Sriskandarajah and Toki Sherbakov and Tomek Korbak and Tomer Kaftan and Tomo Hiratsuka and Tongzhou Wang and Tony Song and Tony Zhao and Troy Peterson and Val Kharitonov and Victoria Chernova and Vineet Kosaraju and Vishal Kuo and Vitchyr Pong and Vivek Verma and Vlad Petrov and Wanning Jiang and Weixing Zhang and Wenda Zhou and Wenlei Xie and Wenting Zhan and Wes McCabe and Will DePue and Will Ellsworth and Wulfie Bain and Wyatt Thompson and Xiangning Chen and Xiangyu Qi and Xin Xiang and Xinwei Shi and Yann Dubois and Yaodong Yu and Yara Khakbaz and Yifan Wu and Yilei Qian and Yin Tat Lee and Yinbo Chen and Yizhen Zhang and Yizhong Xiong and Yonglong Tian and Young Cha and Yu Bai and Yu Yang and Yuan Yuan and Yuanzhi Li and Yufeng Zhang and Yuguang Yang and Yujia Jin and Yun Jiang and Yunyun Wang and Yushi Wang and Yutian Liu and Zach Stubenvoll and Zehao Dou and Zheng Wu and Zhigang Wang},
      year={2026},
      eprint={2601.03267},
      archivePrefix={arXiv},
      primaryClass={cs.CL},
      url={https://arxiv.org/abs/2601.03267}, 
}

@article{Hattie2007,
title ="{T}he {P}ower of {F}eedback",
author={Hattie, John and Timperley, Helen},
  journal={Review of educational research},
  volume={77},
  number={1},
  pages={81--112},
  year={2007},
  publisher={Sage Publications Sage CA: Thousand Oaks, CA}
}

@inproceedings{borges-etal-2024-teach,
    title = "{L}et {M}e {T}each {Y}ou: {P}edagogical {F}oundations of {F}eedback for {L}anguage {M}odels",
    author = {Borges, Beatriz  and
      Tandon, Niket  and
      K{\"a}ser, Tanja  and
      Bosselut, Antoine},
    editor = "Al-Onaizan, Yaser  and
      Bansal, Mohit  and
      Chen, Yun-Nung",
    booktitle = "Proceedings of the 2024 Conference on Empirical Methods in Natural Language Processing",
    month = nov,
    year = "2024",
    address = "Miami, Florida, USA",
    publisher = "Association for Computational Linguistics",
    url = "https://aclanthology.org/2024.emnlp-main.674/",
    doi = "10.18653/v1/2024.emnlp-main.674",
    pages = "12082--12104",
}

@inproceedings{2025.EDM.poster-demo-papers.281,
 address = {Palermo, Italy},
 author = {Kyosuke Takami and Satoshi Sekine and Yusuke Miyao},
 booktitle = {Proceedings of the 18th International Conference on Educational Data Mining},
 doi = {10.5281/zenodo.15870238},
 editor = {Caitlin Mills and Giora Alexandron and Davide Taibi and Giosuè Lo Bosco and Luc Paquette},
 isbn = {978-1-7336736-6-2},
 month = {July},
 pages = {582--585},
 publisher = {International Educational Data Mining Society},
 title = {{E}valuating {L}ocal {LLM}s on {J}apanese {N}ational {U}niversity {E}ntrance {E}xamination {D}ataset in {C}omparison with {S}tudent {P}erformance},
 year = {2025}
}

@article{asendorpf2002puzzle,
  title={The puzzle of personality types},
  author={Jens B. Asendorpf},
  journal={European Journal of Personality},
  year={2002},
  volume={16},
  pages={S1 - S5},
}

@article{wall2019personality,
title = {{Personality profiles and persuasion: An exploratory study investigating the role of the Big-5, Type D personality and the Dark Triad on susceptibility to persuasion}},
journal = {Personality and Individual Differences},
volume = {139},
pages = {69-76},
year = {2019},
issn = {0191-8869},
doi = {https://doi.org/10.1016/j.paid.2018.11.003},
url = {https://www.sciencedirect.com/science/article/pii/S0191886918305865},
author = {Helen J. Wall and Claire C. Campbell and Linda K. Kaye and Andy Levy and Navjot Bhullar},
}

@article{takami2023personality,
  title={{Personality-based tailored explainable recommendation for trustworthy smart learning system in the age of artificial intelligence}},
  author={Takami, Kyosuke and Flanagan, Brendan and Dai, Yiling and Ogata, Hiroaki},
  journal={Smart Learning Environments},
  volume={10},
  number={1},
  pages={65},
  year={2023},
  publisher={Springer},
  doi = {10.1186/s40561-023-00282-6},
  url = {https://doi.org/10.1186/s40561-023-00282-6}
  
}

@article{Valerie2008,
title ={{Focus on Formative Feedback}},
author = {Valerie J. Shute},
journal = {Review of Educational Research},
volume = {78},
number = {1},
year = {2008},
publisher="Sage Publications",
URL = {https://doi.org/10.3102/0034654307313795},
doi = {10.3102/0034654307313795},
pages = {153--189},
}

@article{wisniewski2020power,
  title={{The power of feedback revisited: A meta-analysis of educational feedback research}},
  author={Wisniewski, Benedikt and Zierer, Klaus and Hattie, John},
  journal={Frontiers in psychology},
  volume={10},
  pages={3087},
  year={2020},
  publisher={Frontiers Media SA}
}

@inproceedings{10260740,
  author={Dai, Wei and Lin, Jionghao and Jin, Hua and Li, Tongguang and Tsai, Yi-Shan and Gašević, Dragan and Chen, Guanliang},
  booktitle={2023 IEEE International Conference on Advanced Learning Technologies (ICALT)}, 
  title={{Can Large Language Models Provide Feedback to Students? A Case Study on ChatGPT}}, 
  year={2023},
  volume={},
  number={},
  pages={323-325},
  doi={10.1109/ICALT58122.2023.00100}}

@article{kinder2025effects,
title = {Effects of adaptive feedback generated by a large language model: A case study in teacher education},
journal = {Computers and Education: Artificial Intelligence},
volume = {8},
pages = {100349},
year = {2025},
issn = {2666-920X},
doi = {https://doi.org/10.1016/j.caeai.2024.100349},
url = {https://www.sciencedirect.com/science/article/pii/S2666920X24001528},
author = {Annette Kinder and Fiona J. Briese and Marius Jacobs and Niclas Dern and Niels Glodny and Simon Jacobs and Samuel Leßmann}
}

@article{Escalante2023AIgeneratedFO,
  author = {Escalante, Juan and Pack, Austin and Barrett, Alex},
  title = {{AI-generated feedback on writing: insights into efficacy and ENL student preference}},
  journal = {International Journal of Educational Technology in Higher Education},
  year = {2023},
  volume = {20},
  number = {1},
  pages = {57},
  doi = {10.1186/s41239-023-00425-2},
  url = {https://doi.org/10.1186/s41239-023-00425-2},
  issn = {2365-9440},
}

@article{steiss2024comparing,
title = {{Comparing the quality of human and ChatGPT feedback of students' writing}},
journal = {Learning and Instruction},
volume = {91},
pages = {101894},
year = {2024},
issn = {0959-4752},
doi = {https://doi.org/10.1016/j.learninstruc.2024.101894},
url = {https://www.sciencedirect.com/science/article/pii/S0959475224000215},
author = {Jacob Steiss and Tamara Tate and Steve Graham and Jazmin Cruz and Michael Hebert and Jiali Wang and Youngsun Moon and Waverly Tseng and Mark Warschauer and Carol Booth Olson},
}

@article{asadi2025impact,
  title={{The impact of integrating ChatGPT with teachers' feedback on EFL writing skills}},
  author={Asadi, Marjan and Ebadi, Saman and Mohammadi, Laleh},
  journal={Thinking Skills and Creativity},
  volume={56},
  pages={101766},
  year={2025},
  publisher={Elsevier}
}

@article{mccrae1992introduction,
  title={{An Introduction to the Five-Factor Model and Its Applications}},
  author={McCrae, Robert R and John, Oliver P},
  journal={Journal of personality},
  volume={60},
  number={2},
  pages={175--215},
  year={1992},
  doi = {10.1111/j.1467-6494.1992.tb00970.x}
}

@article{poropat2009meta,
  title={A meta-analysis of the five-factor model of personality and academic performance.},
  author={Poropat, Arthur E},
  journal={Psychological bulletin},
  volume={135},
  number={2},
  pages={322--338},
  year={2009},
 doi = {https://doi.org/10.1037/a0014996}
}

@article{Sharma2025,
  author = {Sharma, Sahil and Mittal, Puneet and Kumar, Mukesh and Bhardwaj, Vivek},
  title = {The Role of Large Language Models in Personalized Learning: A Systematic Review of Educational Impact},
  journal = {Discover Sustainability},
  year = {2025},
  volume = {6},
  number = {1},
  pages = {243},
  doi = {https://doi.org/10.1007/s43621-025-01094-z},
  url = {https://doi.org/10.1007/s43621-025-01094-z},
}

@article{VanLehn2011TheRE,
  title={{The Relative Effectiveness of Human Tutoring, Intelligent Tutoring Systems, and Other Tutoring Systems}},
  author={VanLehn, Kurt},
  journal={Educational psychologist},
  volume={46},
  number={4},
  pages={197--221},
  year={2011},
  publisher={Taylor \& Francis}
}

@inproceedings{Kochmar2020AutomatedPF,
author="Kochmar, Ekaterina
and Vu, Dung Do
and Belfer, Robert
and Gupta, Varun
and Serban, Iulian Vlad
and Pineau, Joelle",
editor="Bittencourt, Ig Ibert
and Cukurova, Mutlu
and Muldner, Kasia
and Luckin, Rose
and Mill{\'a}n, Eva",
title={{Automated Personalized Feedback Improves Learning Gains in An Intelligent Tutoring System}},
booktitle="Artificial Intelligence in Education",
year="2020",
publisher="Springer International Publishing",
address="Cham",
pages="140--146",
isbn="978-3-030-52240-7"
}

@article{Bidjerano2007TheRB,
  title={The relationship between the big-five model of personality and self-regulated learning strategies.},
  author={Temi Bidjerano and David Yun Dai},
  journal={Learning and Individual Differences},
  year={2007},
  volume={17},
  pages={69-81}
}

@article{Aleven2006TowardMT,
  title={{Toward Meta-cognitive Tutoring: A Model of Help Seeking with a Cognitive Tutor}},
  author={Aleven, Vincent and Mclaren, Bruce and Roll, Ido and Koedinger, Kenneth},
  journal={International journal of artificial intelligence in education},
  year={2006},
  volume={16},
  number={2},
  pages={101-128},
  publisher={SAGE Publications Sage UK: London, England}
}

@article{villegas2025adaptive,
  title={{Adaptive intelligent tutoring systems for STEM education: analysis of the learning impact and effectiveness of personalized feedback}},
  author={Villegas-Ch, William and Buenano-Fernandez, Diego and Navarro, Alexandra Maldonado and Mera-Navarrete, Aracely},
  journal={Smart Learning Environments},
  volume={12},
  number={1},
  pages={41},
  year={2025},
  publisher={Springer Nature BV},
  doi = {https://doi.org/10.1186/s40561-025-00389-y}
}

@article{kasneci2023chatgpt,
  title={{ChatGPT for good? On opportunities and challenges of large language models for education}},
  author={Kasneci, Enkelejda and Se{\ss}ler, Kathrin and K{\"u}chemann, Stefan and Bannert, Maria and Dementieva, Daryna and Fischer, Frank and Gasser, Urs and Groh, Georg and G{\"u}nnemann, Stephan and H{\"u}llermeier, Eyke and others},
  journal={Learning and individual differences},
  volume={103},
  pages={102274},
  year={2023},
  publisher={Elsevier}
}

@inproceedings{stamper2024enhancing,
author="Stamper, John
and Xiao, Ruiwei
and Hou, Xinying",
editor="Olney, Andrew M.
and Chounta, Irene-Angelica
and Liu, Zitao
and Santos, Olga C.
and Bittencourt, Ig Ibert",
title={{Enhancing LLM-Based Feedback: Insights from Intelligent Tutoring Systems and the Learning Sciences}},
booktitle="Artificial Intelligence in Education. Posters and Late Breaking Results, Workshops and Tutorials, Industry and Innovation Tracks, Practitioners, Doctoral Consortium and Blue Sky",
year="2024",
publisher="Springer Nature Switzerland",
address="Cham",
pages="32--43",
isbn="978-3-031-64315-6"
}

@article{komarraju2005relationship,
  title={The relationship between the big five personality traits and academic motivation},
  author={Komarraju, Meera and Karau, Steven J},
  journal={Personality and individual differences},
  volume={39},
  number={3},
  pages={557--567},
  year={2005},
  publisher={Elsevier}
}

@article{komarraju2009role,
title = {{Role of the Big Five personality traits in predicting college students' academic motivation and achievement}},
author={Komarraju, Meera and Karau, Steven J and Schmeck, Ronald R},
journal = {Learning and Individual Differences},
volume = {19},
number = {1},
pages = {47-52},
year = {2009},
publisher={Elsevier}
}

@article{Murakami1997BigFive,
  title={{Scale construction of a ``Big Five'' personality inventory (in Japanese)}},
  author={Yoshihiro Murakami and Chieko Murakami},
  journal={The Japanese Journal of Personality},
  volume={6},
  number={1},
  pages={29-39},
  year={1997},
  doi={10.2132/jjpjspp.6.1_29}
}

@inproceedings{jiang-etal-2024-personallm,
    title = {{PersonaLLM: Investigating the Ability of Large Language Models to Express Personality Traits}},
    author = "Jiang, Hang  and
      Zhang, Xiajie  and
      Cao, Xubo  and
      Breazeal, Cynthia  and
      Roy, Deb  and
      Kabbara, Jad",
    editor = "Duh, Kevin  and
      Gomez, Helena  and
      Bethard, Steven",
    booktitle = "Findings of the Association for Computational Linguistics: NAACL 2024",
    month = jun,
    year = "2024",
    address = "Mexico City, Mexico",
    publisher = "Association for Computational Linguistics",
    url = "https://aclanthology.org/2024.findings-naacl.229/",
    doi = "10.18653/v1/2024.findings-naacl.229",
    pages = "3605--3627"
}

@inproceedings{liu-etal-2024-personality,
    title = {{Personality-aware Student Simulation for Conversational Intelligent Tutoring Systems}},
    author = "Liu, Zhengyuan  and
      Yin, Stella Xin  and
      Lin, Geyu  and
      Chen, Nancy F.",
    editor = "Al-Onaizan, Yaser  and
      Bansal, Mohit  and
      Chen, Yun-Nung",
    booktitle = "Proceedings of the 2024 Conference on Empirical Methods in Natural Language Processing",
    month = nov,
    year = "2024",
    address = "Miami, Florida, USA",
    publisher = "Association for Computational Linguistics",
    url = "https://aclanthology.org/2024.emnlp-main.37/",
    doi = "10.18653/v1/2024.emnlp-main.37",
    pages = "626--642"
}

@inproceedings{chu-etal-2025-llm,
    title = {{LLM Agents for Education: Advances and Applications}},
    author = "Chu, Zhendong  and
      Wang, Shen  and
      Xie, Jian  and
      Zhu, Tinghui  and
      Yan, Yibo  and
      Ye, Jingheng  and
      Zhong, Aoxiao  and
      Hu, Xuming  and
      Liang, Jing  and
      Yu, Philip S.  and
      Wen, Qingsong",
    editor = "Christodoulopoulos, Christos  and
      Chakraborty, Tanmoy  and
      Rose, Carolyn  and
      Peng, Violet",
    booktitle = "Findings of the Association for Computational Linguistics: EMNLP 2025",
    month = nov,
    year = "2025",
    address = "Suzhou, China",
    publisher = "Association for Computational Linguistics",
    url = "https://aclanthology.org/2025.findings-emnlp.743/",
    doi = "10.18653/v1/2025.findings-emnlp.743",
    pages = "13782--13810",
    ISBN = "979-8-89176-335-7"
}

@misc{qian2025deanllmtutorsexploring,
      title={{Dean of LLM Tutors: A Framework for Automated Quality Review of AI-generated Feedback}}, 
      author={Keyang Qian and Yixin Cheng and Rui Guan and Wei Dai and Flora Jin and Kaixun Yang and Sadia Nawaz and Zachari Swiecki and Guanliang Chen and Lixiang Yan and Dragan Gašević},
      year={2026},
      eprint={2508.05952},
      archivePrefix={arXiv},
      primaryClass={cs.CY},
      url={https://arxiv.org/abs/2508.05952}, 
}

@inproceedings{chu2026feedevalpedagogicallyalignedevaluation,
      title = {{FeedEval: Pedagogically Aligned Evaluation of {LLM}-Generated Essay Feedback}},
    author = "Chu, SeongYeub  and
      Kim, Jongwoo  and
      Yi, Mun Yong",
    editor = "Liakata, Maria  and
      Moreira, Viviane P.  and
      Zhang, Jiajun  and
      Jurgens, David",
    booktitle = "Findings of the {A}ssociation for {C}omputational {L}inguistics: {ACL} 2026",
    month = jul,
    year = "2026",
    address = "San Diego, California, United States",
    publisher = "Association for Computational Linguistics",
    url = "https://aclanthology.org/2026.findings-acl.615/",
    doi = "10.18653/v1/2026.findings-acl.615",
    pages = "12648--12674",
    ISBN = "979-8-89176-395-1",
}

@inproceedings{lai-chang-2019-tellmewhy,
    title = {{TellMeWhy: Learning to Explain Corrective Feedback for Second Language Learners}},
    author = "Lai, Yi-Huei  and
      Chang, Jason",
    editor = "Pad{\'o}, Sebastian  and
      Huang, Ruihong",
    booktitle = "Proceedings of the 2019 Conference on Empirical Methods in Natural Language Processing and the 9th International Joint Conference on Natural Language Processing (EMNLP-IJCNLP): System Demonstrations",
    month = nov,
    year = "2019",
    address = "Hong Kong, China",
    publisher = "Association for Computational Linguistics",
    url = "https://aclanthology.org/D19-3040/",
    doi = "10.18653/v1/D19-3040",
    pages = "235--240"
}

@inproceedings{sanyal-etal-2025-investigating,
    title = {{Investigating Pedagogical Teacher and Student LLM Agents: Genetic Adaptation Meets Retrieval-Augmented Generation Across Learning Styles}},
    author = "Sanyal, Debdeep  and
      Maiti, Agniva  and
      Maharana, Umakanta  and
      Kumar, Dhruv  and
      Mali, Ankur  and
      Giles, C. Lee  and
      Mandal, Murari",
    editor = "Christodoulopoulos, Christos  and
      Chakraborty, Tanmoy  and
      Rose, Carolyn  and
      Peng, Violet",
    booktitle = "Proceedings of the 2025 Conference on Empirical Methods in Natural Language Processing",
    month = nov,
    year = "2025",
    address = "Suzhou, China",
    publisher = "Association for Computational Linguistics",
    url = "https://aclanthology.org/2025.emnlp-main.675/",
    doi = "10.18653/v1/2025.emnlp-main.675",
    pages = "13348--13389",
    ISBN = "979-8-89176-332-6"
}

@inproceedings{
hu2026simbench,
title={{SimBench: Benchmarking the Ability of Large Language Models to Simulate Human Behaviors}},
author={Tiancheng Hu and Joachim Baumann and Lorenzo Lupo and Nigel Collier and Dirk Hovy and Paul R{\"o}ttger},
booktitle={The Fourteenth International Conference on Learning Representations},
year={2026},
url={https://openreview.net/forum?id=PL51SpN6ZJ}
}

@article{black1998assessment,
  title={Assessment and classroom learning},
  author={Black, Paul and Wiliam, Dylan},
  journal={Assessment in Education: principles, policy \& practice},
  volume={5},
  number={1},
  pages={7--74},
  year={1998},
  publisher={Taylor \& Francis}
}

@article{carless2018development,
  title={{The development of student feedback literacy: Enabling uptake of feedback}},
  author={Carless, David and Boud, David},
  journal={Assessment \& evaluation in higher education},
  volume={43},
  number={8},
  pages={1315--1325},
  year={2018},
  publisher={Taylor \& Francis}
}

@article{lipnevich2016toward,
  title={{Toward a Model of Student Response to Feedback}},
  author={Lipnevich, Anastasiya A and Berg, David AG and Smith, Jeffrey K},
  journal={Handbook of human and social conditions in assessment},
  pages={169--185},
  year={2016}
}

@misc{beale2025revolutionarrivedcurrentstate,
      title={{The Revolution Has Arrived: What the Current State of Large Language Models in Education Implies for the Future}}, 
      author={Russell Beale},
      year={2025},
      eprint={2507.02180},
      archivePrefix={arXiv},
      primaryClass={cs.HC},
      url={https://arxiv.org/abs/2507.02180}, 
}

@misc{addae2026privacydesignframeworklarge,
      title={{A Privacy by Design Framework for Large Language Model-Based Applications for Children}}, 
      author={Diana Addae and Diana Rogachova and Nafiseh Kahani and Masoud Barati and Michael Christensen and Chen Zhou},
      year={2026},
      eprint={2602.17418},
      archivePrefix={arXiv},
      primaryClass={cs.AI},
      url={https://arxiv.org/abs/2602.17418}, 
}
\appendix

\section{Feedback Elements}
\label{app:feedback_element}

~\citet{borges-etal-2024-teach} define ten elements for educational feedback design.
Since our study focuses on text-based feedback, we exclude \textit{Mode}, which represents multimodal feedback, and consider the remaining nine elements and their sub-elements.

\paragraph{Pilot Study}
We generate feedback for the same question using these candidates and qualitatively examine whether each produce distinct educational characteristics. 
We find that several elements and sub-elements generated highly overlapping feedback, making them unsuitable as independent experimental conditions. 
Specifically, we exclude the \textit{Purpose}, \textit{Criteria}, and \textit{Answer Coverage} elements, as well as the \textit{simple} and {technical} sub-elements of \textit{Style}, the \textit{factual} and \textit{improvement} sub-elements of \textit{Valence}, and the \textit{sentence} and \textit{paragraph} sub-elements of \textit{Granularity}, because they  produce feedback that overlap with other conditions.

Therefore, we select elements based on the distinguishability of the generated feedback rather than predefined quantitative criteria. 
We further reduce the number of conditions to five by considering participant workload and the educational importance of the elements.

These five elements are not intended to cover the entire design space of educational feedback. Rather, they are selected to provide sufficiently distinct educational characteristics for exploring how different feedback designs affect learners' revision performance and subjective evaluations.

\section{Participants}
\label{app:participants}
The participants in this study were first-year students enrolled in a public high school in Japan. 
Their ages ranged from 15 to 16 years. 
The school is known for its strong record of students advancing to highly selective universities.
The experiment was conducted as part of a winter vacation assignment for a basic biology course.
Participants were informed in advance that their participation would not affect their academic evaluation and that they would not be disadvantaged if they chose not to participate.
In addition, parental consent was obtained prior to the commencement of the experiment.
The consent form used in the experiment is shown in Table~\ref{tab:consent_form}.
The pre-questionnaire items are shown in Tables~\ref{tb:bigfive_questionnaire}, ~\ref{tb:bigfive_questionnaire2}, and~\ref{tb:questionnaire3}.

\section{Ethical Approval and Data Protection}
\label{app:ethic}

Our study was conducted with approval from our institution's ethics review board.
We also obtained approval from the principal of the participating school, informed consent from the participants' guardians, and assent from the participants themselves.
The study was conducted under the supervision of the students' classroom teachers.
The consent form explicitly stated that participants' personal information would be kept confidential and that the collected data would be anonymized and used for research purposes.
These procedures were implemented to ensure appropriate ethical oversight and data protection throughout the study.

\begin{figure*}[t]
\includegraphics[width=\textwidth]{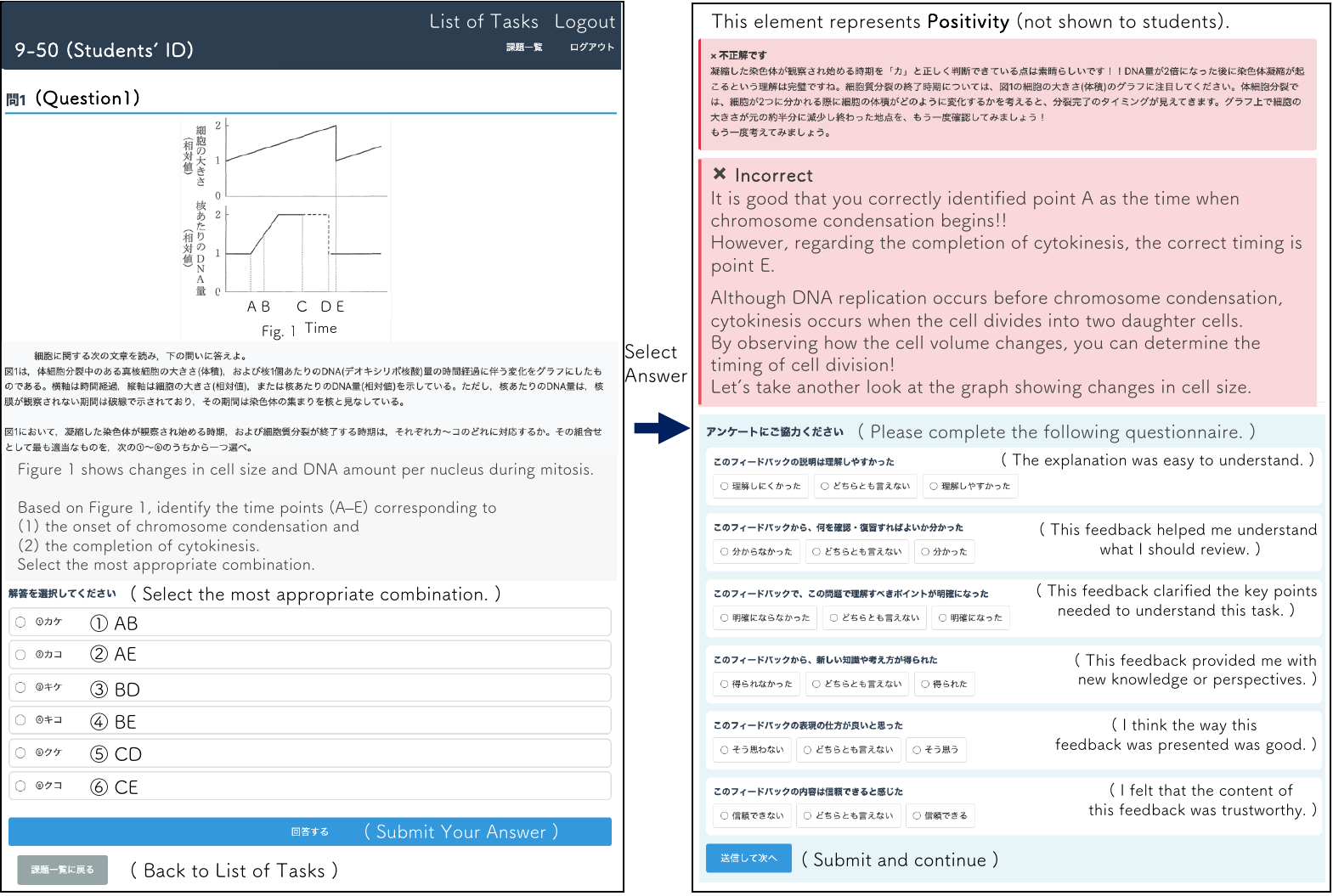}
\caption{
User interface of the system used in this study.
Participants read the question prompt, select an answer option, and submit their response.
The system then provides feedback, which participants evaluate using a three-point scale across six criteria.
This example shows \textit{Positivity}.
} 
\label{fig:UI_example}
\end{figure*}

\section{Question Selection}
\label{app:dataset}
Although the dataset covers the full biology subject, the participants are first-year high school students who have not yet completed all biology units.
We focus on multiple-choice questions for two reasons.
First, multiple-choice questions are widely used in Japanese school education and therefore provide a realistic setting for an empirical classroom study.
Second, because each question has a clearly defined correct answer, they allow us to evaluate the effects of feedback while minimizing ambiguity in judging response correctness.

Because the study was conducted in an authentic educational setting, we also needed to limit the number of questions to reduce participant burden.
The number of questions was determined in consultation with the teacher at the participating school.

We then extract questions corresponding to the basic biology curriculum that the participants had already studied through a two-stage filtering process involving LLMs and domain experts.
First, we use GPT-5~\cite{GPT-5} to identify questions related to the basic biology curriculum from the full set of biology questions.
Specifically, we include key topics from the national curriculum guidelines~\footnote{The national curriculum guidelines:~\url{
https://www.mext.go.jp/component/a_menu/education/micro_detail/__icsFiles/afieldfile/2010/01/29/1282000_6.pdf
}} for basic biology in the prompt and instruct the model to extract questions that correspond to these topics.
As a result, 33 questions are selected from the original set of 101 questions.

Next, two domain experts in biology education conduct further filtering.
They review the questions to ensure that they are appropriate as homework and aligned with the participants' completed curriculum.
Based on their expert review and agreement, we finally select seven questions.

\section{Feedback System}
\label{app:UI}
Figure~\ref{fig:UI_example} illustrates the system's user interface. 
Prior to the experiment, we generated feedback for all possible answer choices using GPT-5~\cite{GPT-5}.
The experimental workflow is as follows:
First, participants enter their student ID and set a password. Next, they select a task from the problem list, which is presented in a randomized order for each participant. 
Upon submitting an answer, participants receive feedback tailored to their specific response; notably, the type of feedback element assigned is also randomized.
Finally, participants evaluate the feedback using a 3-point scale based on the six criteria detailed as follows.
This feedback-loop cycle—encompassing the initial response, feedback reception, and subjective evaluation—continues until the participant reaches the correct answer.

\paragraph{Subjective Evaluation Criteria}
We use a subjective evaluation of the feedback based on six criteria, using a 3-point Likert scale (3: Agree, 2: Neither agree nor disagree, 1: Disagree). 
The evaluation criteria and their respective scales are presented below:

\begin{itemize}
\item \textbf{Guidance for Review}: This feedback helped me understand what I should review.
\item \textbf{Trustworthiness}: I felt that the content of
this feedback was trustworthy.
\item \textbf{New Knowledge}: This feedback provided me with
new knowledge or perspectives.
\item \textbf{Key Points Clarity}: This feedback clarified the key points
needed to understand this task.
\item  \textbf{Ease of Understanding}: The explanation was easy to understand.
\item \textbf{Expression Quality}: I think the way this feedback was presented was good. 
\end{itemize}

\section{Result of Qualitative Analysis}
\label{app:qualitative_analysis}

We conduct a qualitative analysis to verify whether the generated feedback aligns with our educational intentions.

\subsection{Evaluation Criteria}
The authors qualitatively analyze all feedback instances based on the following five dimensions.

\begin{itemize}

\item \textbf{Specificity}:  
This dimension evaluates how concretely the feedback identifies what learners should revise.
Feedback is categorized as either:
\begin{itemize}
    \item \textbf{High}: The feedback clearly specifies incorrect concepts, reasoning steps, or revision targets.
    \item \textbf{Low}: The feedback provides only general comments or abstract encouragement without concrete revision guidance.
\end{itemize}

\item \textbf{Information Density}:  
This dimension evaluates how efficiently the feedback conveys task-relevant information.

\begin{itemize}
    \item \textbf{High}: The feedback conveys task-relevant information concisely and efficiently.
    \item \textbf{Medium}: The feedback provides an appropriate amount of explanation without substantial redundancy or missing information.
    \item \textbf{Low}: The feedback is either overly verbose or lacks sufficient task-relevant information.
\end{itemize}

\item \textbf{Cognitive Scaffolding}:  
This dimension evaluates how the feedback supports learner reasoning processes.
Because a single feedback instance may contain multiple forms of scaffolding, we conduct multi-label annotation using the following categories:
\begin{itemize}
    \item \textbf{Error Identification}: The feedback explicitly points out incorrect reasoning or answers.
    \item \textbf{Conceptual Explanation}: The feedback explains background concepts relevant to the problem.
    \item \textbf{Stepwise Reasoning}: The feedback decomposes causal relations or reasoning processes related to the problem.
    \item \textbf{Knowledge Elaboration}: The feedback provides additional knowledge extending beyond the immediate scope of the problem.
\end{itemize}

\item \textbf{Interaction Framing}:  
This dimension evaluates how the feedback positions the learner socially and pedagogically.
Each feedback instance is categorized into one of the following types:
\begin{itemize}
    \item \textbf{Evaluative}: The feedback mainly focuses on correctness judgments.
    \item \textbf{Encouraging}: The feedback mainly emphasizes praise or emotional support.
    \item \textbf{Expository}: The feedback primarily adopts an explanatory and instructional style.
\end{itemize}

\item \textbf{Educational Reliability}:  
This dimension evaluates whether the feedback preserves productive learner inference without excessively narrowing the answer space.
\begin{itemize}
    \item \textbf{Preserved}: The feedback maintains learner reasoning and avoids revealing the correct answer.
    \item \textbf{Reduced}: The feedback indirectly reveals the answer or substantially reduces the required inference process.
    \item \textbf{Hallucination}:  
This criterion evaluates whether the feedback contains factually incorrect explanations or scientifically inaccurate statements.
\end{itemize}

\end{itemize}

\begin{table}[t]
\centering
\resizebox{\columnwidth}{!}{
\begin{tabular}{lcccc}
\toprule
\textbf{Elements} & \textbf{Error} & \textbf{Conceptual} & \textbf{Stepwise} & \textbf{Knowledge} \\
 & \textbf{Identification} & \textbf{Explanation} & \textbf{Reasoning} & \textbf{Elaboration} \\
\midrule
\textit{Normal} & 28 & 0 & 29 & 7 \\
\rowcolor{LightGray} \textit{Keywords} & 31 & 0 & 31 & 9 \\
\textit{Actionability} & 32 & 0 & 32 & 7 \\
\rowcolor{LightGray} \textit{Novelty} & 30 & 3 & 17 & 36 \\
\textit{Coverage} & 30 & 0 & 30 & 11 \\
\rowcolor{LightGray} \textit{Positivity} & 2 & 0 & 30 & 7 \\
\midrule
\textbf{Total} & 153 & 3 & 169 & 77 \\
\bottomrule
\end{tabular}
}
\caption{Results of Cognitive Scaffolding analysis across feedback elements.}
\label{tb:cognitive_results}
\end{table}
\begin{table}[t]
\centering
\resizebox{\columnwidth}{!}{
\begin{tabular}{lccc}
\toprule
\textbf{Elements} & \textbf{Preserved} & \textbf{Reduced} & \textbf{Hallucination} \\
\midrule
\textit{Normal} & 22 & 7 & 0 \\
\rowcolor{LightGray} \textit{Keywords} & 31 & 2 & 0 \\
\textit{Actionability} & 24 & 9 & 0 \\
\rowcolor{LightGray} \textit{Novelty} & 30 & 0  & 0 \\
\textit{Coverage} & 24 & 9 & 0\\
\rowcolor{LightGray} \textit{Positivity} & 29 & 2 & 0 \\
\midrule
\textbf{Total} & 160 & 29 & 0 \\
\bottomrule
\end{tabular}
}
\caption{Results of Pedagogical Reliability analysis across feedback elements.}
\label{tb:reliability_results}
\end{table}
\begin{table*}[t]
\centering
\resizebox{\textwidth}{!}{
\begin{tabular}{lccccc}
\toprule
\textbf{Elements}
& \textbf{Unfiltered n / Acc (\%)}
& \textbf{Filtered n / Acc (\%)}
& $\Delta$ \textbf{Acc}
& \textbf{Unfiltered $\beta$ (vs. Normal)}
& \textbf{Filtered $\beta$ (vs. Normal)} \\
\midrule
Normal & 198 / 65.7 &	148 / 64.2 &	$-$1.5pt	& ref	& ref \\
\rowcolor{LightGray} Keywords&	204 / 53.0 & 	153 / 58.2 &	$+$5.3pt	& $-$0.530$^{*}$ &	$-$0.254 (n.s.) \\
Actionability	& 202 / 49.0&	143 / 46.9	& $-$2.1pt	&$-$0.688$^{**}$	&$-$0.710$^{**}$\\
\rowcolor{LightGray} Novelty&	213 / 52.1	&213 / 52.1	&$\pm$0&	$-$0.564$^{*}$	&$-$0.499$^{*}$\\
Coverage	&171 / 53.2	&136 / 52.9	&$-$0.3pt	&$-$0.519$^{*}$&	$-$0.466 (marginal, p=.061)\\
\rowcolor{LightGray} Positivity	&208 / 46.6	&185 / 48.6	&$+$2.0pt&	$-$0.783$^{***}$	&$-$0.638$^{**}$ \\

\bottomrule
\end{tabular}
}
\caption{Sensitivity analysis of revision performance using unfiltered and filtered data.
Normal shows the highest revision accuracy in both datasets, while the overall pattern of results remains similar after excluding answer-leaking feedback.
$n$ denotes the number of corresponding responses, and Acc. denotes revision accuracy. $^{}p < .05$, $^{}p < .01$, $^{}p < .001$.}
\label{tb:unfiltered_revision}
\end{table*}
\begin{table*}[t]
\centering
\small
\resizebox{\textwidth}{!}{
\setlength{\tabcolsep}{4pt}

\begin{tabular}{lccccc}
\hline
\textbf{Criterion}
& \textbf{Keywords}
& \textbf{Actionability}
& \textbf{Novelty}
& \textbf{Coverage}
& \textbf{Positivity} \\
\hline

Guidance for Review
& .027 / .040
& .013 / .025
& $-.058$ / $-.064$
& $-.027$ / $-.034$
& $-.127^{**}$ / $-.135^{**}$ \\

\rowcolor{LightGray} Trustworthiness
& .035 / .050
& .025 / .058
& $-.013$ / $-.013$
& .007 / .020
& $-.093^{*}$ / $-.082^{*}$ \\

New Knowledge
& $-.015$ / $-.013$
& $-.061$ / $-.012$
& .053 / .067
& $-.042$ / $-.022$
& $-.123^{**}$ / $-.115^{**}$ \\

\rowcolor{LightGray} Key Points Clarity
& $-.008$ / $-.007$
& $-.009$ / .010
& $-.108^{**}$ / $-.125^{*}$
& $-.047$ / $-.056$
& $-.163^{***}$ / $-.169^{***}$ \\

Ease of Understanding
& .005 / $-.004$
& $-.058$ / $-.041$
& $-.148^{***}$ / $-.162^{***}$
& $-.050$ / $-.042$
& $-.119^{**}$ / $-.125^{**}$ \\

\rowcolor{LightGray} Expression Quality
& .008 / .007
& .001 / .029
& $-.095^{*}$ / $-.110^{**}$
& $-.029$ / $-.019$
& $-.148^{***}$ / $-.158^{***}$ \\

\hline
\end{tabular}
}

\caption{
Sensitivity analysis of subjective evaluations using unfiltered and filtered data.
Normal is the reference condition.
Each cell reports the regression coefficient for the unfiltered / filtered analysis, respectively.
$^{*}p < .05$, $^{**}p < .01$, $^{***}p < .001$.
}
\label{tb:unfiltered_subjective}
\end{table*}
\begin{table}[t]
  \centering
  \setlength{\tabcolsep}{3pt} 
\resizebox{\columnwidth}{!}{
  \begin{tabular}{lcccccc}
    \toprule
    & \multicolumn{2}{c}{} & & & \multicolumn{2}{c}{95\% CI} \\
    \cmidrule(lr){6-7}
    Predictor & $\beta$ & SE & $z$ & $p$ & Lower & Upper \\
    \midrule
    Intercept (Normal)              &  0.584 & 0.181 &  3.227 & $< .001$ &  0.229 &  0.938 \\
    \rowcolor{LightGray}Keywords                          & $-$0.254 & 0.242 & $-$1.048 & $.295$   & $-$0.729 & 0.221 \\
    Actionability     & $-$0.710 & 0.243 & $-$2.921 & $.003$   & $-$1.186 & $-$0.233 \\
    \rowcolor{LightGray}Novelty               & $-$0.499 & 0.221 & $-$2.259 & $.024$   & $-$0.932 & $-$0.066 \\
    Coverage                   & $-$0.466 & 0.248 & $-$1.875 & $.061$   & $-$0.953 & 0.021 \\
    \rowcolor{LightGray}Positivity                        & $-$0.638 & 0.226 & $-$2.825 & $< .005$ & $-$1.080 & $-$0.195 \\
    \bottomrule
  \end{tabular}
  }
\vspace{-1mm}
\caption{Generalized Estimating Equation (GEE) Results for First Revised Accuracy.
    $N = 978$. Reference category: Normal.
  GEE with robust covariance estimator; independence working correlation structure.
  Chi-square test: $\chi^2(5) = 12.611$, $p = .027$.}

  \label{tb:gee_results}
\end{table}
\begin{table}[!t]
  \centering
  \setlength{\tabcolsep}{3pt}
  \resizebox{\columnwidth}{!}{
  \begin{tabular}{lcccccc}
    \toprule
    & & & & & \multicolumn{2}{c}{95\% CI} \\
    \cmidrule(lr){6-7}
    Predictor & $\beta$ & SE & $z$ & $p$ & Lower & Upper \\
    \midrule

    \multicolumn{7}{l}{\textbf{Criterion 1: Guidance for Review}} \\
    \midrule
    Intercept (Normal) & 2.499 &       0.037 &     66.72 &       0.000    &    2.425  &      2.572   \\
    \rowcolor{LightGray}Keywords             & 0.040 &      0.041 &       0.961  &     0.336  &     $-$0.041  &      0.120 \\
    Actionability        & 0.025     &  0.045 &      0.553&       0.580 &      $-$0.063     &   0.112 \\
    \rowcolor{LightGray}Novelty              & $-$0.064 &       0.041   &   $-$1.559   &    0.119   &    $-$0.145  &      0.017 \\
    Coverage             & $-$0.034    &   0.039   &   $-$0.881 &      0.378    &   $-$0.110    &    0.042 \\
    \rowcolor{LightGray}Positivity           & $-$0.135 &      0.045  &    $-$3.037    &   0.002   &    $-$0.223   &    $-$0.048  \\
    \midrule

    \multicolumn{7}{l}{\textbf{Criterion 2: Trustworthiness}} \\
    \midrule
     Intercept (Normal) & 2.521    &  0.035   &  71.55    &  0.000 &      2.452     &  2.590 \\
    \rowcolor{LightGray}Keywords             & 0.050 &    0.032 &     1.559    &  0.119   &   $-$0.013    &   0.113 \\
    Actionability        & 0.058  &    0.034   &   1.715   &   0.086  &    $-$0.008   &    0.123 \\
    \rowcolor{LightGray}Novelty              & $-$0.013   & 0.032     &$-$0.385    &  0.700   &   $-$0.076   &    0.051 \\
    Coverage             & 0.0120   &   0.032   &   0.624  &    0.533  &    $-$0.043   &    0.082 \\
    \rowcolor{LightGray}Positivity           & $-$0.082    &  0.037  &   $-$2.228  &    0.026  &   $-$0.154     & $-$0.010  \\
    \midrule

    \multicolumn{7}{l}{\textbf{Criterion 3: New Knowledge}} \\
    \midrule
    Intercept (Normal) & 2.431    &  0.038 &    64.59 &     0.000   &    2.358    &   2.505    \\
    \rowcolor{LightGray}Keywords             & $-$0.013 &     0.040 &    $-$0.319   &   0.749    &  -0.092    &   0.066  \\
    Actionability        & $-$0.012 &     0.042   &  $-$0.293   &   0.769    &  $-$0.095     &  0.070  \\
    \rowcolor{LightGray}Novelty              & 0.067 &    0.044 &      1.514  &    0.130   &   $-$0.020    &   0.153 \\
    Coverage             & $-$0.022 &     0.042    & $-$0.532  &    0.594  &    $-$0.104   &    0.059  \\
    \rowcolor{LightGray}Positivity           & $-$0.115 &      0.043   &  $-$2.658    &  0.008   &   $-$0.200    &  $-$0.030  \\
    \midrule

    \multicolumn{7}{l}{\textbf{Criterion 4: Key Points Clarity}} \\
    \midrule
    Intercept (Normal) & 2.571  &     0.036   &   71.34     &  0.000   &     2.500   &     2.641     \\
    \rowcolor{LightGray}Keywords             & $-$0.001 &      0.040  &   $-$0.017  &     0.986 &      $-$0.078     &   0.077   \\
    Actionability        & 0.010 &      0.038    &   0.264     &  0.792  &     $-$0.065    &    0.085   \\
    \rowcolor{LightGray}Novelty              & $-$0.125 &    0.040&      $-$3.118 &      0.002  &     $-$0.203   &    -0.046 \\
    Coverage             & $-$0.056 &       0.037  &    $-$1.500 &      0.134 &      $-$0.130    &    0.017    \\
    \rowcolor{LightGray}Positivity           & $-$0.169 &      0.044   &   $-$3.873    &   0.000    &   $-$0.254   &    $-$0.083  \\
    \midrule

    \multicolumn{7}{l}{\textbf{Criterion 5: Ease of Understanding}} \\
    \midrule
    Intercept (Normal) & 2.544 &       0.036  &    70.86 &     0.000   &     2.473    &    2.614 \\
    \rowcolor{LightGray}Keywords             & $-$0.004 &    0.039 &     $-$0.104   &    0.917&       $-$0.081   &     0.073 \\
    Actionability        & $-$0.041 &       0.043  &    $-$0.943 &      0.346 &      $-$0.125    &    0.044 \\
    \rowcolor{LightGray}Novelty              & $-$0.162 &     0.042   &   $-$3.872 &      0.000   &    $-$0.244  &     $-$0.080 \\
    Coverage             & $-$0.042 &       0.041 &     $-$1.041  &     0.298  &     $-$0.122  &     0.037 \\
    \rowcolor{LightGray}Positivity           &$-$0.125 &       0.045&      $-$2.790   &    0.005    &   $-$0.212   &    -0.037 \\
    \midrule

    \multicolumn{7}{l}{\textbf{Criterion 6: Expression Quality}} \\
    \midrule
    Intercept (Normal) & 2.466 &       0.037  &    65.95 &      0.000  &      2.392    &    2.539  \\
    \rowcolor{LightGray}Keywords             & 0.007 &   0.038 &       0.173    &   0.863    &   $-$0.068    &    0.081  \\
    Actionability        & 0.029 &      0.038    &   0.769   &    0.442   &    $-$0.045    &    0.103 \\
    \rowcolor{LightGray}Novelty              & $-$0.110 &    0.040   &   $-$2.731   &    0.006  &     $-$0.189   &    -0.031  \\
    Coverage             & $-$0.019   &    0.038  &    $-$0.503  &     0.615 &      $-$0.094    &    0.055 \\
    \rowcolor{LightGray}Positivity           &  $-$0.158 &      0.046   &   $-$3.470  &     0.001    &   $-$0.247   &    $-$0.069    \\
    \bottomrule
  \end{tabular}
}
\caption{Detailed GEE results for each criterion.
\textit{Positivity} receives significantly lower ratings compared to \textit{Normal} across all criteria. Additionally, \textit{Novelty} shows significant negative effects in three criteria: ``Key Points Clarity'', ``Ease of Understanding'', and ``Expression Quality''.}
\label{tb:gee_subjective_results_all}
\end{table}
\begin{table*}[t]
\centering
\small
\setlength{\tabcolsep}{5pt}

\begin{tabular}{llcc}
\midrule
\textbf{Criterion}
& \textbf{Predictor}
& \textbf{Main OR [95\% CI]}
& \textbf{Sensitivity OR [95\% CI]} \\
\midrule

\multirow{6}{*}{Review Guidance}
& Intercept (Normal) & 1.40 [0.92, 2.13] & 2.10 [1.26, 3.49]$^{**}$ \\
& \cellcolor{LightGray}  Keywords    & \cellcolor{LightGray} 1.18 [0.76, 1.82] & \cellcolor{LightGray}1.02 [0.65, 1.59] \\
& Actionability      & 1.07 [0.67, 1.71] & 1.00 [0.61, 1.63] \\
& \cellcolor{LightGray} Novelty   & \cellcolor{LightGray}0.62 [0.40, 0.95]$^{*}$ & \cellcolor{LightGray}0.59 [0.39, 0.88]$^{*}$ \\
& Coverage           & 1.20 [0.80, 1.79] & 1.21 [0.82, 1.78] \\
& \cellcolor{LightGray} Positivity         & \cellcolor{LightGray}0.86 [0.56, 1.31] & \cellcolor{LightGray}0.83 [0.56, 1.24] \\

\midrule
\multirow{6}{*}{Trustworthiness}
& Intercept (Normal) & 1.76 [1.13, 2.75]$^{*}$ & 2.11 [1.25, 3.56]$^{**}$ \\
& \cellcolor{LightGray}Keywords           &\cellcolor{LightGray} 0.75 [0.52, 1.09] & \cellcolor{LightGray}0.70 [0.48, 1.03] \\
& Actionability      & 0.72 [0.49, 1.08] & 0.70 [0.46, 1.06] \\ 
&\cellcolor{LightGray} Novelty            & \cellcolor{LightGray}0.55 [0.34, 0.87]$^{*}$ & \cellcolor{LightGray} 0.54 [0.35, 0.83]$^{**}$ \\
& Coverage           & 0.80 [0.54, 1.21] & 0.80 [0.54, 1.20] \\
 &\cellcolor{LightGray}  Positivity         &  \cellcolor{LightGray} 0.58 [0.41, 0.83]$^{**}$ &  \cellcolor{LightGray}0.57 [0.41, 0.81]$^{**}$ \\

\midrule
\multirow{6}{*}{New Knowledge}
& Intercept (Normal) & 1.42 [0.98, 2.05] & 1.95 [1.21, 3.14]$^{**}$ \\
 &\cellcolor{LightGray} Keywords           &\cellcolor{LightGray} 0.76 [0.50, 1.14] & \cellcolor{LightGray}0.74 [0.48, 1.15] \\
& Actionability      & 0.86 [0.57, 1.30] & 0.82 [0.54, 1.25] \\
 &\cellcolor{LightGray} Novelty            & \cellcolor{LightGray}0.74 [0.48, 1.13] & \cellcolor{LightGray}0.76 [0.51, 1.14] \\
& Coverage           & 0.84 [0.54, 1.31] & 0.86 [0.55, 1.33] \\
&\cellcolor{LightGray} Positivity         & \cellcolor{LightGray}0.75 [0.51, 1.10] & \cellcolor{LightGray}0.74 [0.51, 1.08] \\

\midrule
\multirow{6}{*}{Key Points Clarity}
& Intercept (Normal) & 1.65 [1.09, 2.51]$^{*}$ & 2.32 [1.34, 3.99]$^{**}$ \\
 &\cellcolor{LightGray} Keywords           & \cellcolor{LightGray} 1.00 [0.66, 1.51] &  \cellcolor{LightGray}0.98 [0.62, 1.53] \\
& Actionability      & 0.97 [0.62, 1.51] & 0.89 [0.56, 1.42] \\
 & \cellcolor{LightGray} Novelty            & \cellcolor{LightGray} 0.60 [0.40, 0.90]$^{*}$ & \cellcolor{LightGray} 0.63 [0.42, 0.94]$^{*}$ \\
& Coverage           & 1.08 [0.74, 1.58] & 1.09 [0.75, 1.58] \\
 &\cellcolor{LightGray}  Positivity         & \cellcolor{LightGray} 0.75 [0.49, 1.13] & \cellcolor{LightGray} 0.74 [0.49, 1.11] \\

\midrule
\multirow{6}{*}{Ease of Understanding}
& Intercept (Normal) & 1.73 [1.16, 2.60]$^{**}$ & 2.48 [1.49, 4.14]$^{***}$ \\
& \cellcolor{LightGray} Keywords           & \cellcolor{LightGray} 0.77 [0.52, 1.15] & \cellcolor{LightGray}0.70 [0.47, 1.05] \\
& Actionability      & 0.80 [0.53, 1.21] & 0.76 [0.49, 1.17] \\
 &\cellcolor{LightGray}  Novelty            &\cellcolor{LightGray}  0.47 [0.30, 0.73]$^{***}$ & \cellcolor{LightGray} 0.48 [0.32, 0.71]$^{***}$ \\
& Coverage           & 0.96 [0.63, 1.46] & 0.99 [0.66, 1.49] \\
& \cellcolor{LightGray} Positivity         & \cellcolor{LightGray} 0.72 [0.48, 1.09] & \cellcolor{LightGray} 0.71 [0.48, 1.05] \\

\midrule
\multirow{6}{*}{Expression Quality}
& Intercept (Normal) & 1.40 [0.92, 2.12] & 1.77 [1.01, 3.10]$^{*}$ \\
 & \cellcolor{LightGray} Keywords           & \cellcolor{LightGray} 0.76 [0.51, 1.14] & \cellcolor{LightGray} 0.72 [0.48, 1.07] \\
& Actionability      & 0.83 [0.54, 1.28] & 0.81 [0.51, 1.27] \\
 & \cellcolor{LightGray} Novelty            &  \cellcolor{LightGray}0.46 [0.29, 0.74]$^{**}$ & \cellcolor{LightGray} 0.47 [0.30, 0.74]$^{**}$ \\
& Coverage           & 0.84 [0.53, 1.33] & 0.87 [0.55, 1.37] \\
 & \cellcolor{LightGray} Positivity         & \cellcolor{LightGray} 0.71 [0.47, 1.06] & \cellcolor{LightGray} 0.69 [0.47, 1.03] \\

\midrule
\end{tabular}

\caption{
Sensitivity analysis controlling for question-level variation.
Odds ratios (ORs) and 95\% confidence intervals are reported for the main logistic GEE model and the sensitivity model including question ID as a fixed effect.
Normal is the reference condition.
$^{*}p < .05$, $^{**}p < .01$, $^{***}p < .001$.
}
\label{tb:question_sensitivity}
\end{table*}

\subsection{Results}
For \textbf{Specificity}, 9.0\% (20/222) of the feedback instances are classified as Low.
All of these cases belong to \textit{Positivity}.
These feedback instances tend to prioritize praise and encouragement while providing limited concrete guidance about what learners should revise.
For \textbf{Information Density}, 12.2\% (27/222) of the feedback instances are classified as Low.
Among them, 13 instances belong to \textit{Positivity} and 14 to \textit{Novelty}.
Similar to the Specificity results, \textit{Positivity} often lacks sufficient task-relevant revision guidance because praise-oriented expressions dominate the feedback.
In contrast, \textit{Novelty} frequently contains rich explanatory content, but the explanations often extend beyond the scope of the target problem, resulting in insufficient guidance for immediate answer revision.

Table~\ref{tb:cognitive_results} presents the results for \textbf{Cognitive Scaffolding}.
\textbf{Error Identification} appears in 68.9\% (153/222) of all feedback instances.
Although most feedback types include explicit identification of learner errors, \textit{Positivity} contains substantially fewer such cases.
\textbf{Conceptual Explanation} appears only three times, all of which belong to \textit{Novelty}.
\textbf{Stepwise Reasoning} appears in 76.1\% (169/222) of the feedback instances and is observed across all feedback types.
\textbf{Knowledge Elaboration} appears in 34.7\% (77/222) of the feedback instances, with nearly half of these cases belonging to \textit{Novelty}.
This result indicates that \textit{Novelty} frequently introduces additional background knowledge beyond the immediate problem scope.

For \textbf{Interaction Framing}, \textbf{Evaluative} and \textbf{Encouraging} styles appear in 4 and 37 instances, respectively.
These correspond primarily to \textit{Normal} and \textit{Positivity}.
In contrast, \textbf{Expository} framing appears in 81.5\% (181/222) of all feedback instances and is distributed relatively evenly across all feedback types except \textit{Positivity}.

Table~\ref{tb:reliability_results} presents the results for \textbf{Educational Reliability}.
72.1\% (160/222) of the feedback instances are classified as \textbf{Preserved}, with \textit{Keywords} showing the largest number of such cases.
In contrast, 13.1\% (29/222) are classified as \textbf{Reduced}, where the feedback indirectly narrows the answer space or reveals the correct answer through detailed explanations.
These cases appear particularly frequently in \textit{Coverage} and \textit{Actionability}.
No instances of factual hallucination are observed.
Appendix~\ref{ssec:example_leaked_feedback} provides examples of leaked feedback instances.

\subsection{Example of Leaked Feedback}
\label{ssec:example_leaked_feedback}
Table~\ref{tb:leaked_feedback} presents a specific instance where the feedback indirectly suggests the correct answer.
The target question requires learners to match specific causes with their effects.
Although the feedback does not explicitly state that the correct answer is \Circled{7}, it provides an explanation of the underlying combinations that makes \Circled{7} the only logical conclusion. 
This type of indirect feedback effectively narrows down the options, resulting in unintended answer leakage.

\section{Sensitivity Analysis for Answer Leakage}
\label{app:sensitivity_leaked}

The objective of this study is to examine how the pedagogical elements of feedback affect learners' revision performance and subjective evaluations.
Answer-leaking feedback, however, may reveal the correct answer through mechanisms unrelated to the pedagogical effect of the feedback element itself.
Including such instances would therefore conflate the effect we aim to measure, namely the effect of feedback design, with an unintended effect of direct answer disclosure. For this reason, excluding answer-leaking feedback is necessary to preserve the validity of the construct being evaluated rather than an arbitrary removal of data.

To assess the impact of this exclusion, we run the same GEE analyses on the unfiltered data and compare the results with those from the filtered data.
For RQ1, as shown in Table~\ref{tb:unfiltered_revision}, Baseline/Normal consistently shows the highest revision accuracy in both datasets, and the overall pattern of results remains similar. The differences for Actionability, Novelty, and Positivity remain significant in both analyses, whereas those for Keywords and Coverage are no longer statistically significant after exclusion. Thus, although the significance of some pairwise comparisons changes, the main conclusion regarding the relative advantage of Baseline/Normal in immediate revision performance remains unchanged.
For RQ2, as shown in Table~\ref{tb:unfiltered_subjective}, the statistical significance of each effect remains unchanged across all the filtered and unfiltered analyses for all six criteria, with coefficient differences of at most 0.05 points on the three-point scale.

Overall, the exclusion is methodologically motivated and does not substantively alter our main conclusions. 
Although selection bias from this exclusion is theoretically possible, the sensitivity analysis suggests that its practical impact on the reported results is limited.

\section{Statistical Analysis of First Revision Accuracy}
\label{app:gee}
To examine the impact of each feedback element on learners' revised accuracy, we perform a Chi-square test and a Generalized Estimating Equation (GEE) analysis.
Table~\ref{tb:gee_results} presents these results.
The chi-square test for overall model fit ($\chi^2(5)=12.611, p=.027$) indicates that revision accuracy differs across feedback elements.

The GEE results show that all feedback elements have negative coefficients relative to \textit{Normal}, although not all differences are statistically significant.
In particular, \textit{Actionability} ($\beta=-0.710, p=.003$), \textit{Positivity} ($\beta=-0.638, p=.005$), and \textit{Novelty} ($\beta=-0.499, p=.024$) are associated with significantly lower revision accuracy.
\textit{Coverage} shows a marginal negative association ($\beta=-0.466, p=.061$), whereas \textit{Keywords} does not significantly differ from \textit{Normal} ($\beta=-0.254, p=.295$).
Notably, \textit{Positivity}, which also receives relatively low subjective evaluations, shows a strong negative association with revision accuracy.
This may suggest that praise-oriented expressions sometimes distract learners from task-relevant corrective guidance.
In addition, hint-oriented feedback designs such as \textit{Actionability} may require learners to perform additional inferential processing before revising their answers. Although such feedback is perceived as useful and supportive, the additional cognitive effort required to interpret and apply the guidance may reduce immediate revision accuracy.

\section{Statistical Analysis of Subjective Evaluation}
\label{app:stat_subject}
\subsection{Correlation Analysis}
First, to examine the coherence of the evaluation metric, we aggregate participants' evaluations of each feedback element across all six criteria and conduct a correlation analysis among the criteria.
Spearman's rank correlation analysis reveals moderate to strong positive correlations among all criteria ($\rho = 0.58$–$0.76$).
This result suggests that learners' subjective evaluations across criteria tend to be correlated.

\subsection{Generalized Estimating Equations (GEE)}

In the original analysis, we treated the three-point subjective scores as continuous outcomes and applied Gaussian GEE.
However, because the three-point scale had limited variance and showed a strong skew toward ``Agree'' responses, we revised the analysis strategy.
We first attempted an ordinal GEE that preserved the three-point structure, but the model failed to converge because of quasi-complete separation.
We therefore dichotomized the ratings into Agree (3) and Not Agree (1--2) and adopted logistic GEE as the primary analysis.

For completeness, Table~\ref{tb:gee_subjective_results_all} reports the results of the Gaussian GEE analysis.
The overall directional patterns are broadly consistent with the logistic GEE results, although some significance patterns differ, particularly for \textit{Positivity}.


\subsection{Sensitivity Analysis Controlling for Question-Level Variation}
To examine whether question-level variation affects the estimated effects of feedback elements, we conduct a sensitivity analysis by adding question ID as a fixed effect to the logistic GEE model.
We use question ID as a fixed effect rather than a crossed random effect because the study includes only seven questions, which is insufficient for stable estimation of item-level variance components.

We compare the main model,
\texttt{score $\sim$ feedback\_type},
with the sensitivity model,
\texttt{score $\sim$ feedback\_type + C(question\_id)}.
As shown in Table~\ref{tb:question_sensitivity}, the estimated effects of the feedback elements remain highly similar across the two models, with only small changes in odds ratios (approximately 0.01--0.05 for most comparisons).
In particular, the main patterns for \textit{Novelty}, \textit{Positivity}, and the other feedback elements remain unchanged, indicating that our conclusions regarding relative differences between feedback designs are robust to adjustment for question-level variation.

The intercept for \textit{Normal} changes more noticeably after adjustment and becomes significant for additional criteria.
This likely reflects differences in baseline evaluation levels across questions, but does not alter the relative comparisons among feedback elements.

\section{Example of Feedback}
\label{sec:feedback_example}
Table~\ref{fb:fb_example_all_type} presents concrete examples of the feedback generated by each method. The examples shown were generated when a learner selected \protect\Circled{2} in response to the correct answer, \protect\Circled{1}.
In \textit{Normal}, the system provides a concise explanation after stating whether the answer is correct or incorrect. 
\textit{Keywords} highlights essential terms by enclosing them in brackets (e.g., 【 】). 
\textit{Actionability} includes specific, actionable instructions, such as ``first organize your reasoning into three steps'' or ``examine all four elements as a checklist:''
\textit{Novelty} incorporates university-level content related to the problem, while \textit{Coverage} provides a comprehensive explanation touching upon all perspectives necessary for answering the question. Finally, \textit{Positivity} employs an encouraging tone, such as ``Focusing on the involvement of the pancreas in the regulation of hypoglycemia is an excellent perspective!''
All feedback are generated in Japanese.

\section{Result of Elbow for Clustering}
\label{app:elbow}
\begin{figure}[t]
\includegraphics[width=\linewidth]{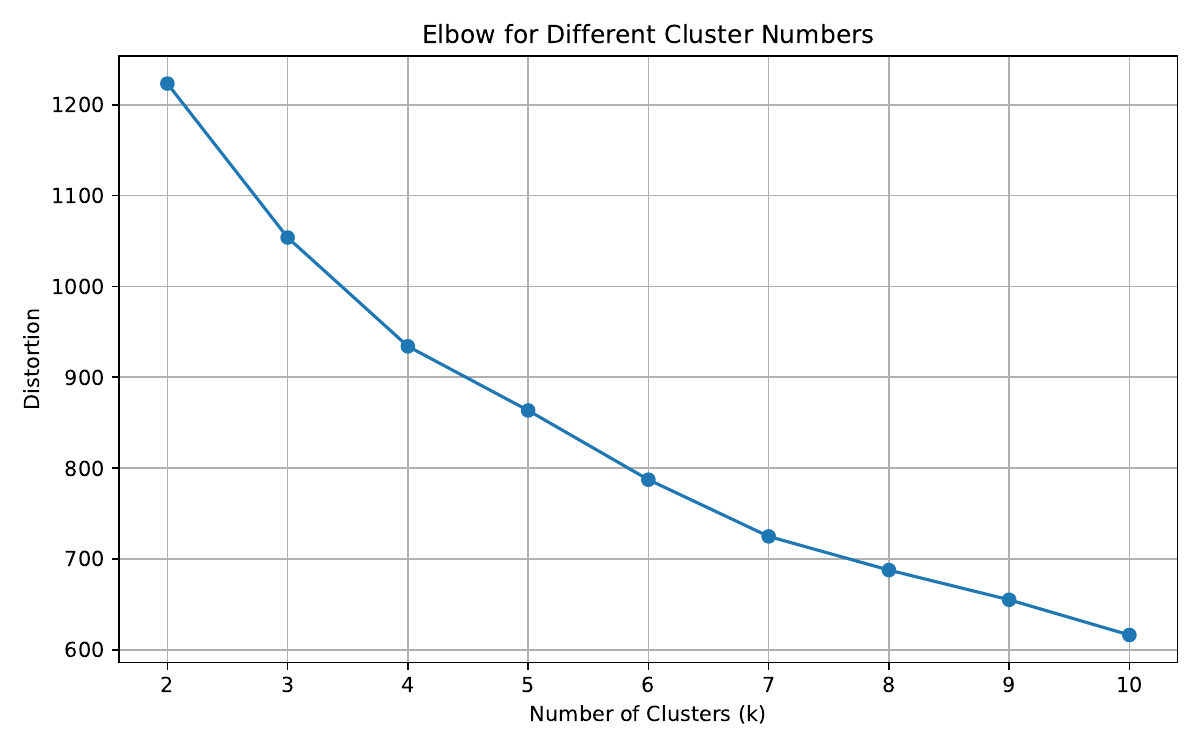}
\caption{
The result of elbow method.
a distinct ``elbow'' or inflection point is observed around $k=3$ and $k=4$, after which the rate of decrease becomes gradual.
} 
\label{fig:elbow}
\end{figure}
To determine the optimal number of clusters for grouping learners' personality traits, we employ the elbow method. Figure~\ref{fig:elbow} presents the results of this analysis. As shown in the figure, a distinct ``elbow'' or inflection point is observed around $k=3$ and $k=4$, after which the rate of decrease becomes gradual.
Based on these results, we consider both three and four as potential cluster counts.
Ultimately, we adopt $k=3$ to maintain consistency with the three personality types widely reported in previous literature~\cite{asendorpf2002puzzle,wall2019personality,takami2023personality}.

\section{Prompt}
\label{sec:prompt}
\begin{figure*}[t]
\includegraphics[width=\textwidth]{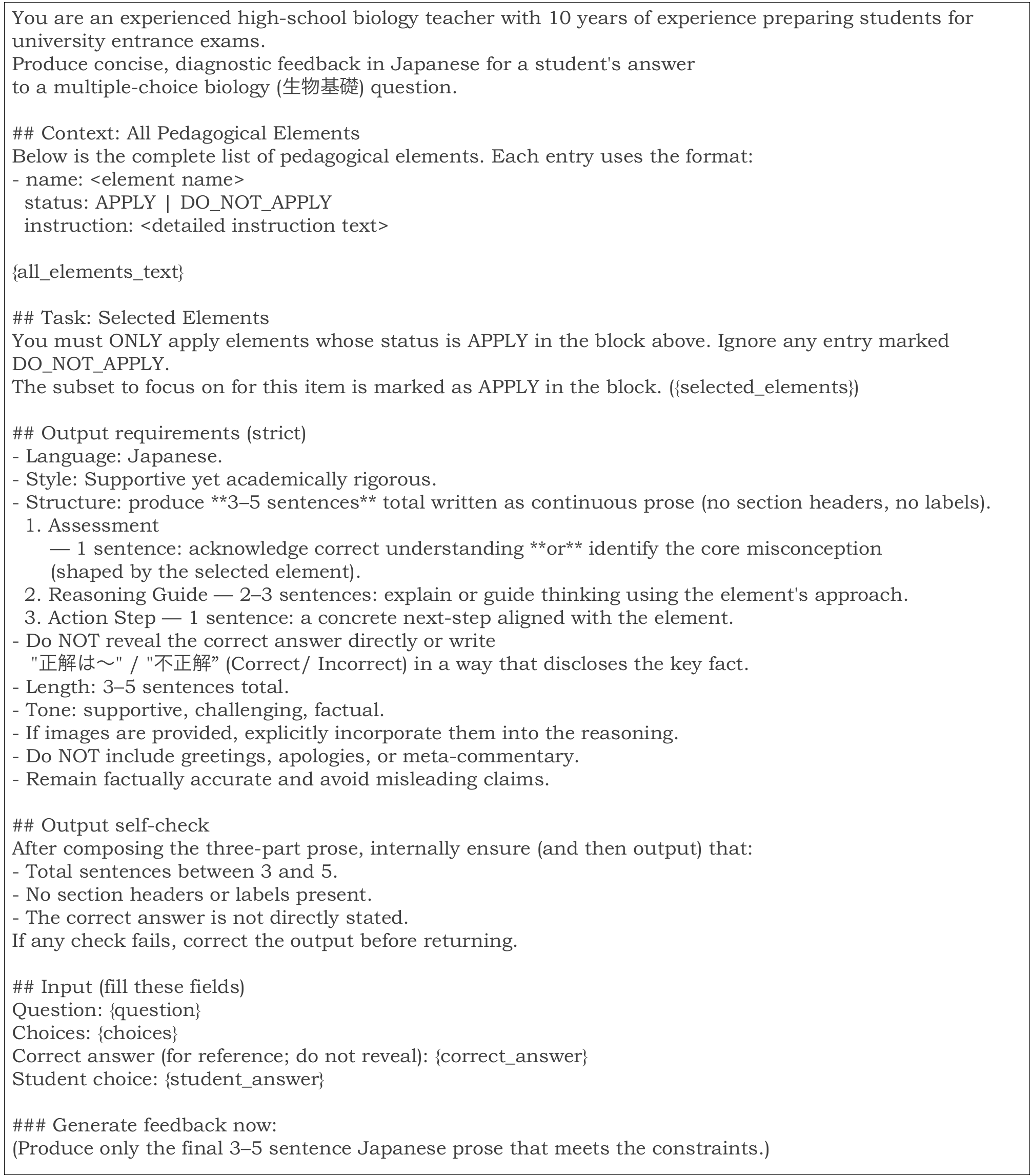}
\caption{
This prompt is overview of the generation process.} 
\label{fig:feedback_generation_prompt}
\end{figure*}

\begin{figure*}[t]
\includegraphics[width=\textwidth]{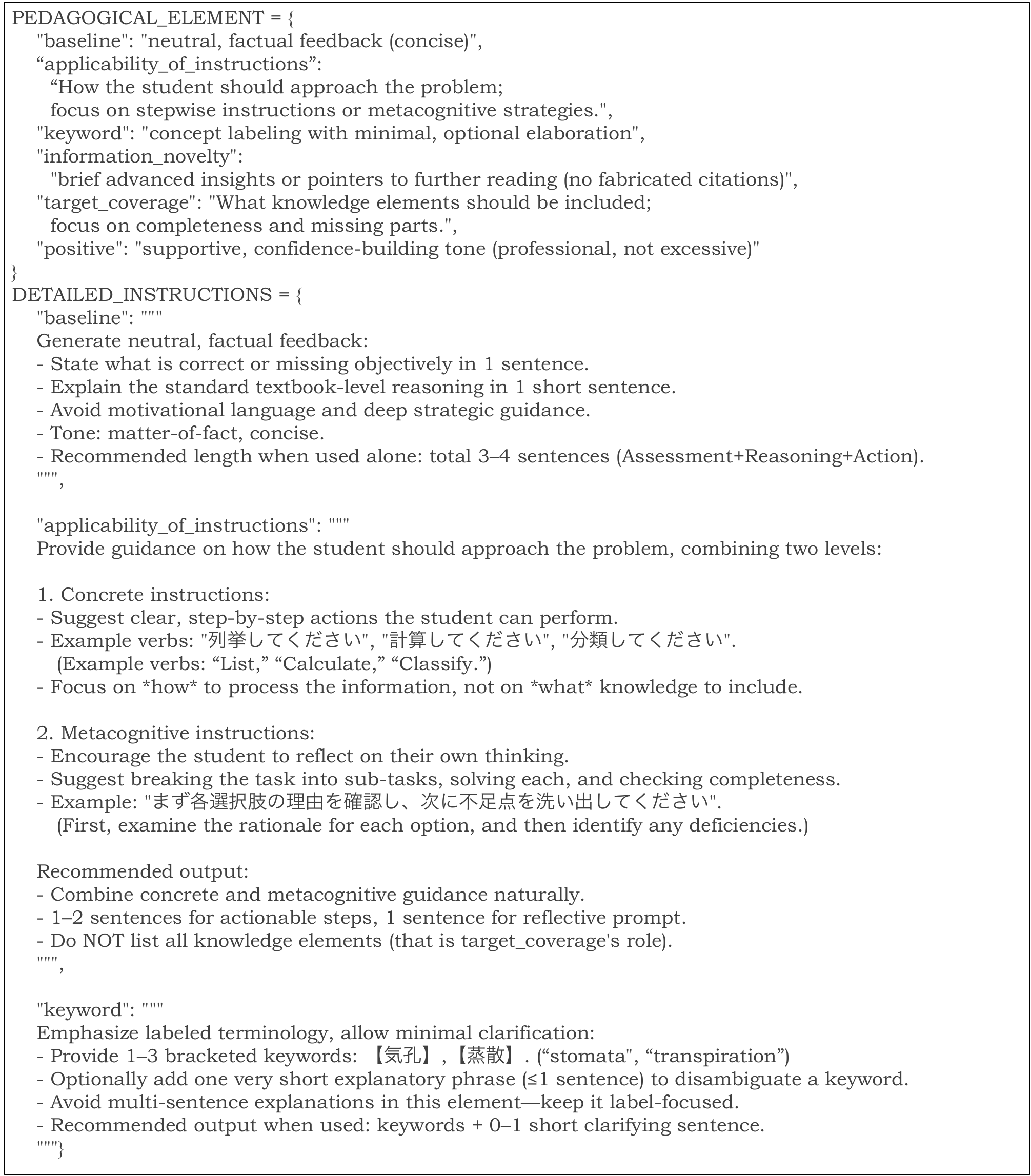}
\caption{
This prompt outlines the specifications for each feedback element, consisting of \texttt{PEDAGOGICAL\_ELEMENT} and \texttt{DETAILED\_INSTRUCTIONS}. 
It presents the \texttt{DETAILED\_INSTRUCTIONS} for \textit{Normal}, \textit{Actionability}, and \textit{Keywords}. The remaining elements are detailed in Figure~\ref{fig:feedback_element_prompt_2}.
}
\label{fig:feedback_element_prompt_1}
\end{figure*}

\begin{figure*}[t]
\includegraphics[width=\textwidth]{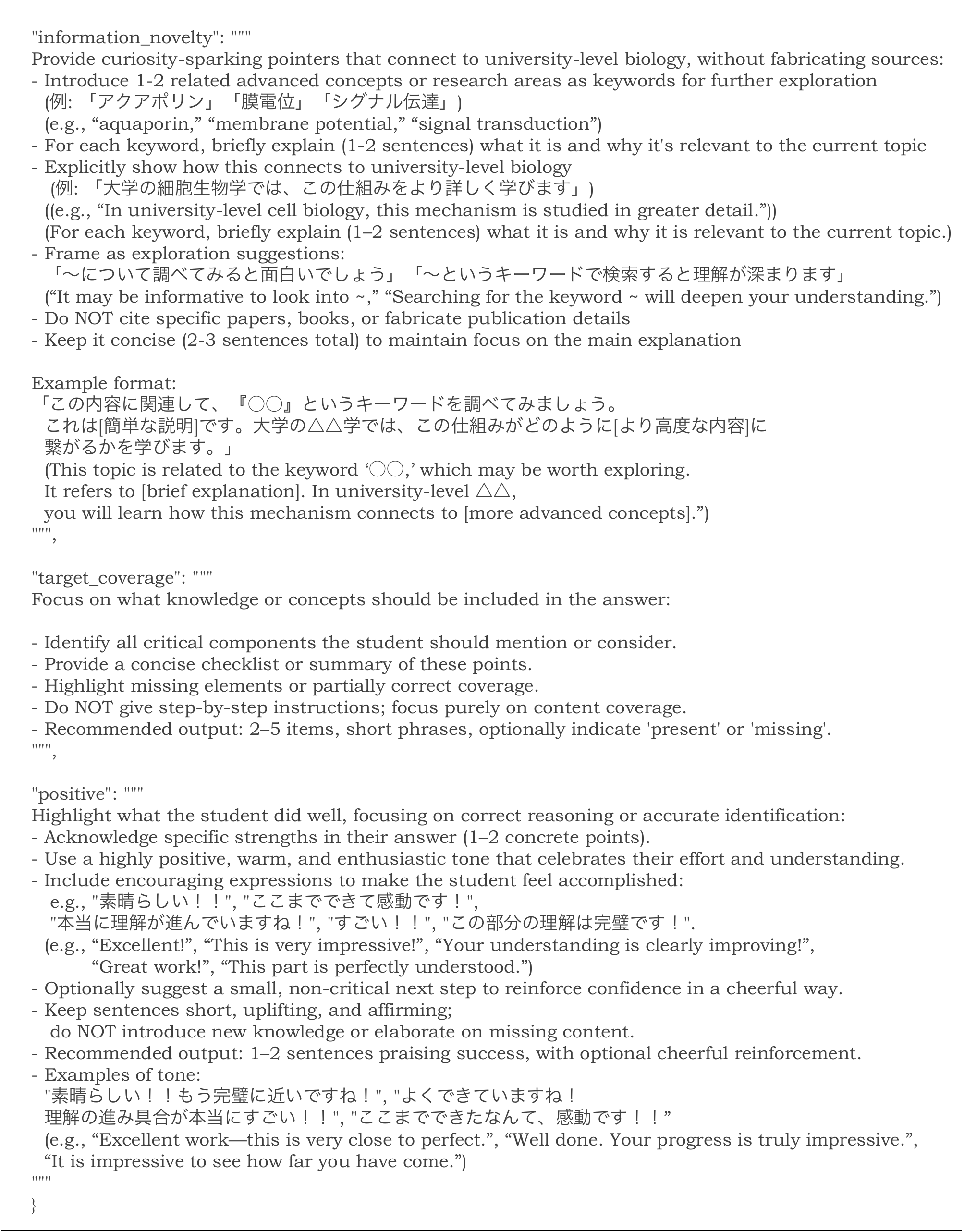}
\caption{
This prompt continues the specifications for each feedback element from Figure~\ref{fig:feedback_element_prompt_1}, specifically presenting the \texttt{DETAILED\_INSTRUCTIONS} for \textit{Novelty}, \textit{Coverage}, and \textit{Positivity}.
}
\label{fig:feedback_element_prompt_2}
\end{figure*}

Figures~\ref{fig:feedback_generation_prompt},~\ref{fig:feedback_element_prompt_1}, and~\ref{fig:feedback_element_prompt_2} present the prompts used for feedback generation. Figure~\ref{fig:feedback_generation_prompt} provides an overview of the generation process, while Figure~\ref{fig:feedback_element_prompt_1} and ~\ref{fig:feedback_element_prompt_2} outline the specific details of each feedback element.

\section{Computational Experiments}
\label{app:computational_experiment}
We select GPT-5~\cite{GPT-5} to generate feedback because some biology questions contain images, which require multimodal understanding, and GPT-5 provides strong overall performance. We use a temperature of 0.1 for feedback generation and do not perform hyperparameter tuning. Generating feedback for all 222 instances required approximately two hours.

\begin{table*}[t]
\centering
\footnotesize
\renewcommand{\arraystretch}{0.9}
\setlength{\extrarowheight}{0.7pt}
\begin{tabular}{>{\raggedright\arraybackslash}m{1.4cm}p{13.5cm}}
\toprule
\textbf{Item} & \textbf{Content} \\
\midrule

Task &

 Answer the following questions regarding ecosystem functions and human health.
Humans live surrounded by various organisms, some of which can be harmful to our health. For example, diseases such as typhoid and dysentery are caused by (Ki: Bacteria), while others like measles and smallpox are caused by (Ku: Viruses). Some organisms enter the human body through food, leading to food poisoning. In addition, there are illnesses such as athlete’s foot caused by (Ke: Fungi).

The processes shown in Figure 1 are essential for maintaining a balanced nutrient cycle within an ecosystem. However, human intervention often disrupts these functions, leading to various environmental issues.
From the options \protect\Circled{1}–\protect\Circled{8} below, select the most appropriate combination of Disruption Causes (u–ka) and the resulting Environmental Problems (b–g).\\&\\&
 (Disruption Causes)\\&
 [u] Discharge of hazardous substances such as heavy metals and DDT.\\&
 [e] Excessive use of chemical fertilizers.\\&
[o] Factory emissions and automobile exhaust gases.\\&
[ka] Reduction in groundwater infiltration due to pavement and the conversion of farmland into residential areas.\\&\\&
(Resulting Problems)\\&
[b] Acid rain causes a loss of biodiversity in flora and fauna.\\&
[c] Substances accumulate through the food chain (biomagnification), causing harm to the human body.\\&
[d] Ozone holes form, leading to global warming. (Note: Scientifically, ozone depletion and global warming are distinct, but this reflects the original text.)\\&
[e] Increased productivity in rivers and lakes leads to eutrophication.\\&
[f] Spring water dries up, resulting in a loss of biodiversity.\\&[g] Forest reduction leads to the progression of desertification.

\\[0.3em]

\midrule
Choices &
\Circled{1} bcdf, \Circled{2} bcfg, \Circled{3} bgcd, \Circled{4}  bgde, \Circled{5} cbfe, \Circled{6} cdeg, \Circled{7} cebf
\Circled{8}  cedg\\
[0.3em]

\midrule
\texttt{Coverage} (Choice: \protect\Circled{2}) &
You attempt to match each cause with its corresponding result, but you must verify which specific problem each factor (u–ka) triggers. Specifically, focus on these direct correlations: heavy metals and DDT lead to biomagnification [c], chemical fertilizers cause eutrophication [e], factory and automobile emissions result in acid rain [b], and reduced groundwater infiltration leads to the depletion of spring water [f]. While Option ④ includes desertification and ozone holes, these do not directly link to the causes (u–ka) specified in the problem. Please re-examine the correspondences (u $\to$ c, e $\to$ e, o $\to$ b, ka $\to$ f) and select the correct combination.
\\[0.3em]

\rowcolor{LightGray}
\texttt{Baseline} (Choice: \protect\Circled{5})  &
You are experiencing some confusion in correctly linking several causes and their results. Please systematically organize the causal relationships based on their environmental mechanisms: heavy metals and DDT trigger biomagnification (u $\to$ c), excessive chemical fertilizer use leads to eutrophication (e $\to$ e), factory and automobile emissions cause acid rain (o $\to$ b), and reduced groundwater infiltration results in the depletion of spring water (ka $\to$ f). Verify exactly how each disruption factor affects specific parts of the ecosystem, and carefully re-evaluate the sequence of problem combinations corresponding to the order of causes (u, e, o, ka).
\\

\bottomrule
\end{tabular}
\caption{Examples of leaked feedback.
These examples illustrate feedback generated when a learner selects \protect\Circled{2} (\textit{Coverage}) or \protect\Circled{5} (\textit{Baseline}) instead of the correct answer, \protect\Circled{7}. The problem requires matching specific causes with their corresponding issues. Although the feedback does not explicitly state that \protect\Circled{7} is the correct answer, it indirectly reveals the solution. By considering the causal combinations presented in the feedback, the learner inevitably identifies \protect\Circled{7} as the only logical choice.
The task and feedback shown here are translated from the original Japanese.
}
\label{tb:leaked_feedback}
\end{table*}

\begin{table*}[t]
\centering
\footnotesize
\renewcommand{\arraystretch}{1.0}
\setlength{\extrarowheight}{0.7pt}

\begin{tabular}{>{\raggedright\arraybackslash}m{1.4cm}p{13.5cm}}
\toprule
\textbf{Item} & \textbf{Content} \\
\midrule

Task &
{\fontsize{8}{10}\selectfont 
Glucose is an essential energy source for the cells that constitute the human body, and it is continuously supplied to all cells via the bloodstream. ...
Therefore, to ensure stable cellular activity, the human body is equipped with a regulatory mechanism that maintains blood glucose levels (the concentration of glucose in the blood) within a constant range.
Diabetes mellitus is a disease in which this regulation of blood glucose fails, resulting in the excretion of glucose in the urine. To diagnose diabetes and determine appropriate treatment strategies, there is a test in which 75 g of glucose is administered in a fasting state, and blood glucose levels and blood insulin concentrations are measured before and after ingestion. This test is known as the oral glucose tolerance test (OGTT).
With regard to the mechanism described in brackets, select the most appropriate regulatory pathway that operates during hypoglycemia from among options \Circled{1} to \Circled{4}.}

{\tiny グルコース（ブドウ糖）は，私たちの体を構成する細胞にとって重要なエネルギー源であり，血液によってすべての細胞に常に供給されている．...
そこで，安定した細胞活動を保証するため，私たちの体には【血糖値（血液中のグルコース濃度）を一定に保つ血糖調節の仕組み】が備わっている．
糖尿病は，この血糖調節がうまくいかなくなり，尿中にグルコースが排出される病気である．
糖尿病の診断と治療方針を決めるため，空腹時に75gのグルコースを飲み，その前後で血糖値や血液中のインスリン濃度などを調べる検査がある．
これを糖負荷試験という．
【】に関して，低血糖時にはたらく調節の反応経路として最も適当なものを，次の\Circled{1}～\Circled{4}のうちから一つ選べ．}
\\[0.3em]

\midrule
Choices &
{\fontsize{8}{10}\selectfont \Circled{1} Diencephalon → Pituitary → Adrenal gland → Glucocorticoids} {\tiny
\Circled{1} 間脳 → 脳下垂体 → 副腎 → 糖質コルチコイド} \\ &
{\fontsize{8}{10}\selectfont \Circled{2} Diencephalon → Parasympathetic NS → Pancreas → Adrenaline} {\tiny \Circled{2} 間脳 → 副交感神経 → 膵臓 → アドレナリン} \\ &
{\fontsize{8}{10}\selectfont \Circled{3} Spinal cord → Sympathetic NS → Adrenal gland → Glucagon} {\tiny \Circled{3} 脊髄 → 交感神経 → 副腎 → グルカゴン}  \\ &
{\fontsize{8}{10}\selectfont \Circled{4} Spinal cord → Parasympathetic NS → Pancreas → Insulin}
{\tiny \Circled{4} 脊髄 → 副交感神経 → 膵臓 → インスリン} \\[0.3em]

\midrule
\texttt{Normal} &
{\fontsize{8}{10}\selectfont There is an error in the combination of the parasympathetic nervous system and the hormone secreted by the pancreas in the regulatory pathway during hypoglycemia.
During hypoglycemia, it is necessary to increase blood glucose levels; therefore, the sympathetic nervous system becomes dominant, and hyperglycemic hormones are secreted from the pancreas and adrenal glands. ...
Review the secretion sites of each hormone (pancreatic $\alpha$-cells and $\beta$-cells, adrenal medulla and adrenal cortex) and organize whether each hormone acts to increase or decrease blood glucose levels.}

{\tiny 低血糖時の調節経路について，副交感神経と膵臓から分泌されるホルモンの組み合わせに誤りがあります．
低血糖時には血糖値を上昇させる必要があるため，交感神経系が優位になり，膵臓や副腎から血糖上昇ホルモンが分泌されます．...
各ホルモンの分泌部位（膵臓の$\alpha$細胞・$\beta$細胞，副腎髄質・副腎皮質）と，それぞれが血糖値を上昇させるか低下させるかを整理し直してください．}
\\[0.3em]

\rowcolor{LightGray}
\texttt{Keywords} &
{\fontsize{8}{10}\selectfont
During hypoglycemia, it is necessary to increase blood glucose levels; therefore, it should be reaffirmed that the sympathetic nervous system, rather than the parasympathetic nervous system, is activated. ...
Furthermore, under the emergency condition of hypoglycemia, \textbf{``the diencephalon (specifically the hypothalamus)''} serves as the central regulator, integratively controlling both the endocrine and autonomic nervous systems. As a result, \textbf{``glucocorticoids are secreted from the adrenal''} cortex and adrenaline from the adrenal medulla. ...}
{\tiny 低血糖時には血糖値を上げる必要があるため，副交感神経ではなく交感神経系が活性化される点を再確認しましょう．... また，低血糖という緊急事態では\textbf{【間脳（視床下部）】}が中心となって内分泌系と自律神経系の両方を統合的に調節し，\textbf{【副腎皮質からの糖質コルチコイド】}や副腎髄質からのアドレナリンが分泌されます．...}
\\[0.3em]

{\scriptsize\texttt{Actionability}}
 &
 {\fontsize{8}{10}\selectfont When considering the regulatory pathway during hypoglycemia, \textbf{first organize your reasoning into three steps: (1) which organ detects the decrease in blood glucose levels, (2) which nervous system is involved, and (3) which hormone is secreted and from which organ.} ... 
For each option, examine all four elements as a checklist: the starting point of the pathway (diencephalon or spinal cord), the type of nervous system involved, the hormone-secreting organ, and the final hormone, and determine whether they are all appropriate for the hypoglycemic response.}

{\tiny \textbf{低血糖時の調節経路を考える際は，まず「どの器官が血糖低下を感知するか」「どの神経系が関与するか」「どのホルモンがどの器官から分泌されるか」という3つのステップに分けて整理してください．} ... 各選択肢について，経路の出発点（間脳か脊髄か），神経系の種類，ホルモン分泌器官，最終的なホルモンの4要素すべてが低血糖応答として正しいかどうかをチェックリストのように検証してください．}
\\[0.3em]

\rowcolor{LightGray}
\texttt{Novelty} &
 {\fontsize{8}{10}\selectfont ... In relation to this topic, it may be informative to examine the keyword ``HPA axis in the stress response.'' The HPA axis (hypothalamus–pituitary–adrenal cortex axis) is an important pathway that increases blood glucose levels by secreting glucocorticoids under stress conditions such as hypoglycemia.
\textbf{In university-level endocrinology, one studies in greater detail how this axis integrates various stress responses and regulates gene expression via glucocorticoid receptors.}...}

{\tiny ... この内容に関連して，「ストレス応答におけるHPA軸」というキーワードを調べてみると面白いでしょう．HPA軸(視床下部-下垂体-副腎皮質軸)は，低血糖などのストレス時に糖質コルチコイドを分泌し血糖値を上げる重要な経路です．\textbf{大学の内分泌学では，この軸がどのように多様なストレス応答を統合し，グルココルチコイド受容体を介して遺伝子発現を調節するかをより詳しく学びます．}...}
\\[0.3em]

\texttt{Coverage} &
{\fontsize{8}{10}\selectfont In considering the regulatory pathways during hypoglycemia, it is necessary to systematically organize the combinations of the nervous systems involved and the hormones released. ...
\textbf{When evaluating the options, verify the validity of each pathway from the following three perspectives: (1) the distinction between the adrenal medulla and adrenal cortex, (2) the difference in function between the sympathetic and parasympathetic nervous systems, and (3) whether each hormone (glucocorticoids, adrenaline, glucagon, and insulin) acts to increase or decrease blood glucose levels.}
}

{\tiny 低血糖時の調節経路について，関与する神経系とホルモンの組み合わせを整理する必要があります．... \textbf{選択肢を検討する際には，\Circled{1}副腎髄質と副腎皮質の違い，\Circled{2}交感神経と副交感神経の作用の違い，\Circled{3}各ホルモン（糖質コルチコイド，アドレナリン，グルカゴン，インスリン）が血糖値を上昇させるか下降させるか，の三つの観点から各経路の妥当性を確認してください．}}
\\[0.3em]

\rowcolor{LightGray}
\texttt{Positivity} &
{\fontsize{8}{10}\selectfont
\textbf{Focusing on the involvement of the pancreas in the regulation of hypoglycemia is an excellent perspective! The pancreas plays a crucial role as a central organ in blood glucose regulation, and recognizing its importance is highly commendable!} ... 
 By systematically matching each hormone with its effect on blood glucose levels (whether increasing or decreasing) and identifying the associated neural and endocrine pathways. ...}

{\tiny 低血糖時の調節に膵臓が関与する点に着目したのは素晴らしい視点です！！膵臓は血糖調節の中心的な器官として非常に重要な役割を果たしているので，その点に気づけたことは本当にすごいことですね！... 低血糖という緊急時に体がどのような指令系統とホルモンで対応するか，特にストレス応答の経路を整理してみてください．...}
\\

\bottomrule
\end{tabular}
\caption{Examples of feedback generated by each element. 
These tasks in this study are sourced from Japanese university entrance examinations. 
All tasks and feedback are originally presented in Japanese.
The examples shown illustrate the feedback generated when a learner selects \protect\Circled{2} instead of the correct answer, \protect\Circled{1}.}
\label{fb:fb_example_all_type}
\end{table*}

\begin{table*}[t]
\centering
\small
\begin{tabular}{p{0.47\linewidth} p{0.47\linewidth}}
\toprule
\textbf{English} & \textbf{Japanese} \\
\midrule

\textbf{Title:} Evaluating the Effects of Generative AI-Based Personalized Feedback on Learning Motivation in High-School Biology &
\textbf{タイトル：} 生物基礎における生成AIによる個別フィードバックの学習意欲向上効果の検証 \\

\textbf{Research Objective:} This study investigates how personalized feedback generated by generative AI for high-school biology questions influences learners' motivation and understanding. The findings will contribute to research and development of generative AI technologies. &
\textbf{研究の目的：}
この研究は、高校生を対象に、生物基礎の問題に対して生成AIが作る個別フィードバック（コメント）が、どのような要素を持つと学習意欲や理解に効果的かを調査し、得られた知見等を広く生成AIの研究開発や普及に役立てることを目的としています。 \\

Generative AI produces feedback using multiple methods, and participants evaluate the quality of the feedback. By analyzing these evaluations, we identify which feedback characteristics are more effective for learning motivation and understanding. &
複数の方法で生成AIがフィードバックを生成し、その効果を高校生の皆さんに評価してもらいます。その評価結果を分析することで、どの要素のフィードバックがより学習意欲や理解に効果的かを明らかにします。 \\

\textbf{Procedure:}
Participants complete the following activities as part of their winter vacation assignment:
(1) submit answers through the system,
(2) review and evaluate AI-generated feedback,
(3) answer pre/post questionnaires,
(4) complete confirmation tests,
and (5) allow collection of learning logs such as study time and progress. &
\textbf{実施手順：}
冬休みの課題の一環として、以下の手順で活動を行っていただきます。
(1) 課題の解答をシステム上で提出
(2) 生成AIが作成したフィードバック（コメント）の確認・評価
(3) 事前アンケート・事後アンケートへの回答
(4) 確認テストの実施
(5) 学習ログ（取り組み時間、進捗、課題達成率など）の収集 \\

\textbf{Data Usage and Consent:}
Collected data such as answers, questionnaires, and learning logs are gathered as part of educational activities. Whether these data are used for research purposes depends on participant consent. &
\textbf{データの取り扱いと同意について：}
この活動で得られるデータ（課題解答、アンケート、学習ログなど）は、教育活動の一環として収集されます。それらの収集されるデータを研究目的で利用するかどうかについては、皆さんの同意に基づいて決定されます。 \\

If participants agree, anonymized data may be used for academic purposes such as research publications and conference presentations. If participants do not agree, they can still complete the assignment and receive AI-generated feedback without disadvantage. &
「同意する」を選択した場合：
収集したデータを研究・学会発表・論文執筆などの学術目的に使用します。
データは匿名化され、個人が特定されることはありません。

「同意しない」を選択した場合：
研究利用は行いませんが、通常どおりAIによるフィードバックを受け取り、課題を行うことができます。 \\

\textbf{Confidentiality:}
All collected data are securely managed and analyzed after anonymization. No personally identifiable information will be disclosed in publications. &
\textbf{機密保持：}
収集したデータは厳重に管理され、研究成果を公表する際にも個人が特定されることはありません。 \\

\textbf{Voluntary Participation:}
Participation in this study is completely voluntary, and participants may withdraw at any time without penalty. &
\textbf{自由意志による参加：}
本研究への参加は完全に任意です。途中で中止する場合も、不利益やペナルティは一切ありません。 \\

\bottomrule
\end{tabular}
\caption{Consent Form Used in the Empirical Experiment.}
\label{tab:consent_form}
\end{table*}
\begin{table*}[t]
\centering
\small
\begin{tabular}{p{0.47\linewidth} p{0.47\linewidth}}
\toprule
\textbf{English} & \textbf{Japanese} \\
\midrule

\textbf{Title:} Evaluating the Effects of Generative AI-Based Personalized Feedback on Learning Motivation in High-School Biology: Pre-questionnaire &
\textbf{タイトル：} 生物基礎における生成AIによる個別フィードバックの学習意欲向上効果の検証　事前アンケート \\

\textbf{Research Title:} Evaluating the Effects of Generative AI-Based Personalized Feedback on Learning Motivation in High-School Biology &
\textbf{研究タイトル：} 生物基礎における生成AIによる個別フィードバックの学習意欲向上効果の検証 \\

This is a pre-questionnaire for evaluating the effects of personalized feedback generated by generative AI on learning motivation. Please select the option that best matches your thoughts. The questionnaire takes approximately 15 minutes and consists of 70 yes/no questions. &
これは、生成AIによる個別フィードバックの学習意欲向上効果の検証のための事前アンケートです。
以下の質問の中から、最も近い選択肢を選択してください。

所要時間はおよそ15分です。
以下の質問に対して、「はい」または「いいえ」で回答してください。
質問は70問あります。 \\

\midrule

I often act without carefully considering problems. &
問題を綿密に検討しないで、実行に移すことが多い \\

I tend to be somewhat lazy. &
どちらかというと、怠惰な方です。 \\

Compared with others, I am talkative. &
他の人と比べると話好きです。 \\

I tend to be quiet and inconspicuous. &
どちらかというと、地味で目立たない方です。 \\

I am considerate of others. &
思いやりがある方です。 \\

I cannot completely trust even close friends. &
親しい仲間でも、本当に信用することはできません。 \\

I can think ahead about the future. &
将来のことを見通すことができる方です。 \\

I tend to worry about trivial things. &
どうでもいいことを、気に病む傾向があります。 \\

I do not get tired easily. &
疲れやすくはありません。 \\

I tend to make decisions impulsively. &
軽率に物事を決めたり、行動してしまいます。 \\

I have a lively personality. &
どちらかというと、にぎやかな性格です。 \\

I work energetically on tasks and studies. &
仕事や勉強には精力的に取り組みます。 \\

I worry about things unnecessarily. &
自分で悩む必要のないことまで心配してしまうのは確かです。 \\

I am not good at speaking in front of people. &
人前で話すのは苦手です。 \\

I try to be kind to everyone. &
誰にでも親切にするように心がけています。 \\

I am not very anxious. &
あまり心配性ではありません。 \\

I believe I am not particularly nervous. &
他の人と同様に、神経質ではないと信じています。 \\

I tend to become emotionally upset easily. &
どちらかというと、気持ちが動揺しやすい。 \\

I actively socialize with others. &
積極的に人と付き合う方です。 \\

I do not care much about how others see me. &
特に人前を気にする方ではありません。 \\

I tend to do things thoroughly. &
どちらかというと、徹底的にやる方です。 \\

I often become confused when facing difficult problems. &
難しい問題にぶつかると、頭が混乱することが多い。 \\

I tend to be shy. &
どちらかというと、引っ込み思案です。 \\

Compared with others, I worry a lot. &
他の人と比べると、あれこれ悩んだり、思いわずらったりする方です。 \\

I try to cooperate with group decisions. &
みんなで決めたことは、できるだけ協力しようと思います。 \\

I tend to overcomplicate things. &
物事を難しく考えがちです。 \\

I tend to get bored easily. &
どちらかというと、飽きっぽい方です。 \\

I quickly want to give up when things do not go well. &
物事がうまくいかないと、すぐに投げ出したくなります。 \\

Something is always bothering me. &
いつも何かが気がかりです。 \\

I know words from many different fields. &
いろいろな分野の言葉を知っています。 \\

\bottomrule
\end{tabular}
\caption{Big Five personality trait questionnaire items used in this study~\cite{Murakami1997BigFive}. The questionnaire consists of 70 items.}
\label{tb:bigfive_questionnaire}
\end{table*}
\begin{table*}[t]
\centering
\small
\begin{tabular}{p{0.47\linewidth} p{0.47\linewidth}}
\toprule
\textbf{English} & \textbf{Japanese} \\
\midrule






I tend to be cautious of others' kindness. &
人から親切にされると、何か下心がありそうで警戒しがちです。 \\

I am not good at analyzing problems. &
問題を分析するのは苦手な方です。 \\

I have given up on things because I thought I lacked ability. &
自分にはそれをする力がないと思って、あきらめてしまったことが何回かあります。 \\

I think I could greatly contribute to society if given the opportunity. &
機会さえあれば、大いに世の中に役立つことができるのにと思います。 \\

I tend to have a quiet personality. &
どちらかというと、おとなしい性格です。 \\

I often quit things halfway. &
何かに取り組んでも、中途半端でやめてしまうことが多い。 \\

I do not strongly assert my opinions. &
あまり自分の意見を主張しない方です。 \\

I make friends easily. &
他の人と同じように、すぐに友達ができる方です。 \\

I certainly lack confidence. &
私は確かに自信に欠けています。 \\

I am better than others at finding common patterns among problems. &
いろいろな問題や事柄から共通した性質を見つけ出すのは、ほかの人より得意です。 \\

I think I could become a good leader if given the opportunity. &
機会さえ与えられれば、皆のよいリーダーになれると思います。 \\

I am an important person. &
私は重要人物です。 \\

Most acquaintances like me. &
ほとんどの知人から好かれています。 \\

I am always worried about something. &
いつも気がかりなことがあって、落ち着きません。 \\

I do not want to cooperate if it disadvantages me. &
みんなで決めたことでも、自分に不利になる場合は協力したくありません。 \\

I am knowledgeable about many things. &
ひろく物事を知っている方です。 \\

I rarely think of unconventional methods. &
いつもと違ったやり方を、なかなか思いつきません。 \\

I tend to be warm-hearted. &
どちらかというと、人情があつい方です。 \\

Even sincere work does not benefit people much. &
誠実に仕事をしても、あまり得にはなりません。 \\

I am full of confidence. &
自信に満ちあふれています。 \\

I think logically. &
筋道を立てて物事を考える方です。 \\

I become flustered easily. &
すぐに、まごまごします。 \\

Compared with others, I act actively. &
ほかの人と比べると活発に行動する方です。 \\

I can remain calm even in upsetting situations. &
大抵の人が動揺するような場合でも、落ち着いて対処することができます。 \\

I work toward clear goals with appropriate methods. &
はっきりとした目標を持って、適切なやり方で取り組みます。 \\

I brood over things. &
くよくよ考え込みます。 \\

People say I am energetic. &
元気がよいと言われます。 \\

I am very poor at speaking in front of classmates. &
学校ではクラスの人たちの前で話すのがひどく苦手でした（です）。 \\

I tend to think more sophisticatedly than others. &
ほかの人より洗練された考え方をする方です。 \\

I tend to be taciturn. &
どちらかというと、無口です。 \\

Compared with others, I can grasp the essence of things. &
ほかの人と比べると、物事の本質が見抜ける方です。 \\

I worry even about trivial matters. &
こまごまとしたことまで気になってしまいます。 \\

I believe people’s words often hide ulterior motives. &
人の言葉には裏があるので、そのまま信じない方が良いと思います。 \\

I tend to lose perseverance easily. &
どちらかというと、三日坊主で、根気がない方です。 \\

I try to think from others’ perspectives. &
いつも人の立場になって考えるように心がけています。 \\

I often become nervous and irritated. &
緊張してイライラすることがよくあります。 \\

I usually make detailed plans before trips. &
旅行などでは、あらかじめ細かく計画を立てることが多い。 \\

I help others even if it is troublesome. &
人助けのためなら、やっかいなことでもやります。 \\

Talking to strangers is exhausting for me. &
初対面の人と話をするのは骨が折れるものです。 \\

I like taking care of children and elderly people. &
子供や老人の世話をするのが好きです。 \\

\bottomrule
\end{tabular}
\caption{Remaining items from Table~\ref{tb:bigfive_questionnaire} of the Big Five personality trait questionnaire used in this study~\cite{Murakami1997BigFive}.}
\label{tb:bigfive_questionnaire2}
\end{table*}
\begin{table*}[t]
\centering
\small
\label{tab:pre_questionnaire_additional}
\begin{tabular}{p{0.47\linewidth} p{0.47\linewidth}}
\toprule
\textbf{English} & \textbf{Japanese} \\
\midrule

The following questions ask about your prior understanding, learning experience, and learning motivation regarding biology and AI-assisted learning. Please select the option that best matches yourself. &
これから取り組む生物基礎の学習やAIを使った学習に関して、皆さんの事前の理解度・学習経験・学習意欲についてお聞きします。質問はすべて、あなた自身に最もあてはまるものを選んでください。 \\

\midrule

I am good at basic biology. &
生物基礎は得意な教科である \\

1: Very poor \hspace{1em} 5: Very good &
1: とても苦手 \hspace{1em} 5: とても得意 \\

\midrule

How frequently do you use generative AI (e.g., ChatGPT, Gemini, Copilot) in daily life? &
あなたは、日常生活の中で生成AI（例：ChatGPT、Gemini、Copilotなど）をどの程度の頻度で利用していますか \\

Never &
全く利用していない \\

Occasionally (less than once a month) &
ときどき利用する（1ヶ月に1回未満） \\

Sometimes (a few times a month) &
ある程度利用する（月に数回） \\

Frequently (a few times a week) &
よく利用する（週に数回） \\

Very frequently (almost every day) &
非常に頻繁に利用する（ほぼ毎日） \\

\midrule

I have used generative AI for studying before. &
これまでに生成AIを使って勉強したことがある \\

Yes &
勉強したことがある \\

No &
勉強したことがない \\

\midrule

How much do you trust information or answers generated by AI? &
生成AIが作った情報や答えをどのくらい信頼しているか \\

1: Do not trust at all \hspace{1em} 5: Trust very much &
1: 全く信頼できない \hspace{1em} 5: とても信頼できる \\

\midrule

I think AI-generated feedback is useful. &
AIによるフィードバックは役に立つと思う \\

1: Not useful at all \hspace{1em} 5: Very useful &
1: 全く役に立たないと思う \hspace{1em} 5: とても役に立つと思う \\

\bottomrule
\end{tabular}
\caption{Additional Pre-questionnaire Items Used in the Empirical Experiment.}
\label{tb:questionnaire3}
\end{table*}
\end{document}